\documentclass[conference]{IEEEtran}
\IEEEoverridecommandlockouts
\usepackage{amsmath,amsfonts}
\usepackage{array}
\usepackage{textcomp}
\usepackage{stfloats}
\usepackage{url}
\usepackage{amsmath, amssymb, amsthm}
\usepackage{verbatim}
\usepackage{graphicx}
\usepackage{amssymb}
\usepackage{hyperref}
\usepackage{graphicx}
\usepackage{amsmath}
\usepackage{longtable}
\usepackage{algorithm} 
\usepackage{algpseudocode} 
\usepackage{mathrsfs}
\usepackage{subcaption}
\usepackage{mathtools}
\usepackage{pifont}
\usepackage{color}
\usepackage{lineno}
\usepackage{graphicx}  
\usepackage{makecell} 
\usepackage{pdflscape}
\usepackage{adjustbox}
\usepackage[utf8]{inputenc}
\usepackage{tabularx}
\usepackage{blindtext}
\usepackage{longtable}
\usepackage{lscape}
\usepackage{amsthm}
\usepackage{graphicx}
\usepackage{booktabs}
\usepackage{amsmath, amssymb}
\usepackage{amsthm}
\usepackage[numbers]{natbib}
\usepackage{tikz}
\usepackage{siunitx}
\usepackage{mathrsfs}
\usepackage{multirow}
\usetikzlibrary{shapes,arrows}
\usepackage{xcolor}
\usepackage{setspace}
\usepackage{notoccite} 
\usepackage{lscape} 
\usepackage{mwe}
\usepackage{booktabs}
\usepackage{amsthm}

\theoremstyle{definition}

\theoremstyle{definition}

\newcommand{\RNum}[1]{\lowercase\expandafter{\romannumeral #1\relax}}
\newcommand{\RNumU}[1]{\uppercase\expandafter{\romannumeral #1\relax}}
\usepackage[numbers]{natbib}

\def\BibTeX{{\rm B\kern-.05em{\sc i\kern-.025em b}\kern-.08em
    T\kern-.1667em\lower.7ex\hbox{E}\kern-.125emX}}
\begin{document}

\title{RoBell-RVFL: A Robust Generalized Bell Random Vector Functional Link Network}

\author{\IEEEauthorblockN{A. Rahaman}
\IEEEauthorblockA{\textit{Department of Mathematics} \\
\textit{Indian Institute of Technology Indore}\\
Indore, India \\
phd2401141001@iiti.ac.in}
\and
\IEEEauthorblockN{A. Quadir}
\IEEEauthorblockA{\textit{Department of Mathematics} \\
\textit{Indian Institute of Technology Indore}\\
Indore, India \\
mscphd2207141002@iiti.ac.in}
\and
\IEEEauthorblockN{M. Tanveer}
\IEEEauthorblockA{\textit{Department of Mathematics} \\
\textit{Indian Institute of Technology Indore}\\
Indore, India \\
mtanveer@iiti.ac.in}
}

\maketitle
\begin{abstract}
The dominance of majority classes in real-world datasets poses a fundamental challenge to randomized neural networks, often biasing decision boundaries and overlooking critical minority samples. Existing remedies, such as synthetic minority over-sampling (SMOTE) and class-weighted loss functions, primarily address class proportions while neglecting intra-class distribution, making them vulnerable to label noise and outliers. In this paper, we propose \textbf{RoBell-RVFL}, a robust and lightweight \emph{quality-aware} generalized bell random vector functional link network that redefines how randomized models handle class imbalance and noisy data. RoBell-RVFL employs a dual-strategy, sample-level weighting mechanism that strictly preserves minority class information using unit weights, while adaptively regulating the influence of majority class samples through a probability-weighted generalized bell (gbell) membership function in a kernel-induced feature space. This design effectively suppresses noisy, boundary, and outlier samples within the majority class, enabling the network to learn from informative samples rather than merely abundant ones. By explicitly incorporating local class probability and class distribution information into the learning process, RoBell-RVFL achieves adaptive control over sample contributions without sacrificing the closed-form learning efficiency of RVFL networks. Extensive evaluations on UCI and KEEL benchmark datasets, along with robustness tests under up to 40\% label noise, demonstrate that RoBell-RVFL consistently and significantly outperforms recent state-of-the-art RVFL variants. The results indicate that adaptive, quality-aware sample weighting is essential for robust RVFL learning, rendering conventional global weighting schemes ineffective in noisy and imbalanced environments.
\end{abstract}

\section{Introduction and Motivation}

\IEEEPARstart{T}{he} random vector functional link (RVFL) network~\cite{PAO1994163, quadir2026garfln} is a prominent randomized neural network (RdNN) model that has gained considerable attention due to its architectural simplicity and high computational efficiency. The defining characteristic of RVFL lies in its learning paradigm, in which the weights connecting the input and hidden layers are randomly initialized and kept fixed throughout training, while only the output-layer weights are analytically learned via a closed-form solution using ridge regression~\cite{suganthan2021origins}. By eliminating iterative weight updates and gradient-based optimization, RVFL enables extremely fast training with low computational overhead, while ensuring stable and reproducible learning behavior.

Moreover, RVFL adopts a shallow architecture with a single hidden layer and direct links between the input and output layers, enabling the preservation of raw input information and improved generalization. Owing to its closed-form learning of output weights and compact network structure with few trainable parameters, RVFL achieves rapid training and low computational complexity \cite{MALIK2023110377}. Guided by principles such as Occam’s razor and probably approximately correct (PAC) learning theory~\cite{kearns1994introduction}, RVFL can be viewed as a lightweight yet effective alternative to conventional artificial neural networks.

Motivated by its architectural simplicity and analytically tractable learning framework, numerous RVFL-based variants have been proposed to further enhance predictive accuracy and robustness across diverse application scenarios~\cite{husmeier1999random, tanveer2025robust}. Owing to its direct input-to-output connections and efficient information flow, RVFL exhibits strong generalization capability with high learning efficiency.

Recent studies have sought to enhance the representational power, robustness, and adaptability of RVFL networks through diverse architectural refinements and learning strategies. For instance, total-var-RVFL and class-var-RVFL~\cite{10.1007/978-3-030-63823-8_48} exploit statistical dispersion in both the input and randomized feature spaces to enable more effective estimation of output-layer weights. The RVFL+ model~\cite{ZHANG202094}, developed under the learning using privileged information (LUPI) paradigm, leverages auxiliary training-only information to guide learning and improve convergence and generalization; its kernelized extension \cite{quadir2025twin, quadir2025trkm}, KRVFL+, further enhances nonlinear modeling capability through the kernel trick \cite{quadir2024one}. Another prominent research direction integrates fuzzy logic into the RVFL framework to better handle uncertainty and imprecision. The neuro-fuzzy RVFL (NF-RVFL)~\cite{10416391} embeds fuzzy inference within RVFL, improving robustness and interpretability.

More recently, complex-valued RVFL (CRVFL)~\cite{LIU2025112682} and its augmented extension (ACRVFL) have been introduced to address challenges related to data representation, incremental learning, and computational efficiency. Additionally, granular and graph-based RVFL variants, including GB-RVFL, GE-GB-RVFL~\cite{SAJID2025111142}, and the multiview graph-based RVFL (GRVFL-MV)~\cite{TANVEER2025121947}, have demonstrated improved discriminative capability by incorporating geometric and multiview information. Despite these advances, a fundamental challenge persists in real-world pattern classification: the presence of severely imbalanced data, often compounded by noise and outliers. In the real world, from face recognition systems to medical diagnostics ~\cite{hanmandlu2014new}, data is rarely perfect. We are frequently confronted with a majority class that overwhelms a minority class, often cluttered with noise and misleading outliers. This isn't just an inconvenience; it is a fundamental flaw in how standard models learn. Conventional RVFL networks are democratic to a fault, they treat every sample equally. When faced with a massive majority class, the network effectively ignores the minority, creating a biased decision boundary that renders the model useless for the cases that often matter most.

To date, existing approaches for addressing class imbalance can be broadly categorized into two groups: data-level and algorithm-level methods. Data-level approaches attempt to rebalance the training set through majority-class undersampling or minority-class oversampling techniques, such as SMOTE~\cite{laurikkala2001improving}. Although widely adopted, these methods suffer from inherent limitations: undersampling may discard informative samples, while oversampling can introduce redundancy and amplify noise~\cite{rezvani2023broad}. 

Algorithm-level approaches, on the other hand, modify the learning mechanism itself. Within the RVFL family, several notable efforts have been reported. CSWRVFL~\cite{sahani2019fpga} addresses power quality disturbance classification, and GE-IFRVFL-CIL~\cite{ganaie2024graph} incorporates class-dependent weighting while preserving the topological structure of the data.

Despite these advances, a critical limitation remains. Existing methods primarily address the \emph{quantity} of class imbalance, while largely overlooking the \emph{quality} of training samples. In particular, they lack mechanisms to identify and suppress noisy or outlier samples within the majority class, while simultaneously preserving the discriminative information of minority samples \cite{11228582}. Consequently, the majority-class noise continues to bias decision boundaries in imbalanced learning scenarios.

To address this gap, we propose the {Robust Generalized Bell RVFL (RoBell-RVFL)}, which performs fine-grained, sample-level weighting rather than coarse class-level reweighting. The core innovation of RoBell-RVFL lies in a probability-weighted generalized bell (gbell) membership function~\cite{10439648}, which integrates spatial distribution information with local class probability. Specifically, minority class samples are assigned unit membership values to ensure full information preservation, whereas majority class samples are adaptively weighted based on their gbell membership and local class probability. Samples located near cluster centers retain higher influence, while boundary, noisy, and outlier samples are effectively suppressed. By coupling this adaptive membership mechanism with a principled class-imbalance weighting scheme~\cite{rezvani2022intuitionistic, QUADIR2024128258}, RoBell-RVFL enables the network to learn from informative samples rather than merely abundant ones.

The main contributions of this work are summarized as follows:
\begin{enumerate}
    \item {Adaptive Sample-Level Framework:} We propose a dual-strategy learning framework that preserves minority class samples using unit weights, while adaptively reducing the influence of majority class samples according to their distribution characteristics.
    
    \item {Probability-Weighted Noise Suppression:} A class probability–guided generalized bell (gbell) membership function is introduced to distinguish informative samples from noise and outliers, thereby preventing majority-class dominance in the learned decision boundary.

    \item {Extensive Empirical Validation:} Comprehensive experiments on UCI and KEEL benchmark datasets, including evaluations under up to 40\% label noise, demonstrate that RoBell-RVFL consistently outperforms state-of-the-art RVFL-based and imbalance-learning methods.
\end{enumerate}

The remainder of this paper is organized as follows. Section~\ref{Sec:Proposed_work} presents the mathematical formulation of RoBell-RVFL and its class-imbalance weighting mechanism. Related work, experimental setup, and experimental results on UCI and KEEL  without label noise are reported in Section S.I., S.IV,  and S.VII. of Supplementary Material. Finally, Section~\ref{Sec:Conclusion} concludes the paper.

\section{Notation}
Consider the training dataset, denoted as \(({a}_i, {b}_i)\), where \({a}_i \in \mathbb{R}^{1 \times B}\) and \({b}_i \in \mathbb{R}^{1 \times D}\), for \(i = 1, 2, \ldots, A\). Here, \(B\) represents the number of features in each input sample, and \(D\) is the total number of classes in the dataset, while \(A\) indicates the total number of training samples. In the context of classification, \({b}_i\) refers to a one-hot encoded vector that signifies the class label corresponding to the input sample \({a}_i\). We define the collections of all input samples and target values as follows:
\[
\scalebox{0.8}{$
Z = \begin{bmatrix}  
{a}_1^t & {a}_2^t & \dots & {a}_M^t  
\end{bmatrix}^t \in \mathbb{R}^{A \times B}, \quad  
Z_{out} = \begin{bmatrix}  
{b}_1^t & {b}_2^t & \dots & {b}_M^t  
\end{bmatrix}^t \in \mathbb{R}^{A \times E}
$}\text{,}\]

where the operator \((\cdot)^t\) represents the transpose. These matrices, \({Z}\) and \({Z_{out}}\), encapsulate all input samples and their corresponding target values, respectively. $D$ is the number of hidden nodes.

\section{Proposed Method}
\label{Sec:Proposed_work}
Assume that the training dataset consists of $Z$ labeled samples given by $\{(a_1, b_1), (a_2, b_2), \ldots, (a_A, b_A)\}$, where $a_i \in \mathbb{R}^B$ denotes the $B$-dimensional feature vector of the $i^{th}$ sample and
$b_i \in \{+1,-1\}$ represents its corresponding class label. Let $h_1 \in \mathbb{R}^{A_1 \times B}$
and $h_2 \in \mathbb{R}^{A_2 \times B}$ denote the matrices formed by the samples belonging to the
positive $(+1)$ and negative $(-1)$ classes, respectively. Without loss of generality, it is assumed that the positive class constitutes the majority class
($A_1 \ge A_2$), while the negative class represents the minority class. A sample $a_i$ belonging
to the majority class is denoted by $a_i^{m}$, whereas a sample belonging to the minority
class is denoted by $a_{i_{m}}$.

In the proposed framework, explicit use of the imbalance ratio is avoided at the sample-weight definition stage. Instead, the effect of class imbalance is addressed through adaptive sample
weighting and class-dependent regularization in the learning phase. Furthermore, the data samples
are implicitly mapped into a higher-dimensional feature space through a kernel-induced mapping,
denoted by $K(\cdot)$, which enables effective modeling of nonlinear class distributions.

\subsection{Determination of Membership Value}

The generalized bell (gbell) function is a flexible fuzzy membership function widely used in fuzzy systems due to its smooth controllability~\cite{10439648}. In the proposed RoBell-RVFL framework, the gbell function assigns adaptive membership values to training samples based on their distribution in a kernel-induced feature space. Specifically, it down-weights outliers and boundary samples from the majority class while preserving full membership for minority class samples to avoid information loss \cite{quadir2024granular}. Unlike conventional distance-based formulations, gbell memberships are computed using a kernel function, enabling implicit mapping into a higher-dimensional space. Let $K(\cdot,\cdot)$ denote a Gaussian kernel defined as $K(a_i,a_j) = \exp\left(-\frac{\|a_i-a_j\|^2}{\xi^2}\right),$
where $\xi$ is the kernel width parameter. The parameter $\xi$ controls the width of the Gaussian kernel and determines the scale of similarity in the kernel-induced feature space. Smaller values of $\xi$ emphasize local neighborhood structures, whereas larger values lead to smoother similarity distributions.

For the positive (majority) and negative (minority) class samples, the kernel-induced radii are computed as
\begin{equation}
r_i^{(+)} =
\sqrt{
1
- 2\,\frac{1}{t_1}\sum_{a_j \in T_1} K(a_i,a_j)
+ \frac{1}{t_1^2}\sum_{a_p,a_q \in T_1} K(a_p,a_q)
},
\end{equation}

\begin{equation}
r_i^{(-)} =
\sqrt{
1
- 2\,\frac{1}{t_2}\sum_{a_j \in T_2} K(a_i,a_j)
+ \frac{1}{t_2^2}\sum_{a_p,a_q \in T_2} K(a_p,a_q)
}.\label{6}
\end{equation}

Let
\begin{equation}
r_{\max}^{(+)} = \max_i r_i^{(+)}, \qquad
r_{\max}^{(-)} = \max_i r_i^{(-)}.
\label{7}
\end{equation}

The gbell membership functions for the positive and negative class samples are defined as
\begin{equation}
\xi_{\text{pw}}^{(+)}(a_i)
=
\frac{1}
{1 +
\left(
\frac{r_i^{(+)}}{c_1}
\right)^{2d_1}},
\qquad
c_1=d_1=\rho\, r_{\max}^{(+)},
\label{8}
\end{equation}

\begin{equation}
\xi_{\text{pw}}^{(-)}(a_i)
=
\frac{1}
{1 +
\left(
\frac{r_i^{(-)}}{c_2}
\right)^{2d_2}},
\qquad
c_2=d_2=\rho\, r_{\max}^{(-)},
\label{9}
\end{equation}
where $\rho>0$ is a scaling parameter controlling the effective width and steepness of the gbell function.

\subsubsection{Determination of Class Probability}

To further reduce the influence of noise and ambiguous samples during training, a class probability value is assigned to each training sample. This probability reflects the local class consistency of a sample within a kernel-induced neighborhood and is used to suppress noisy and mislabeled samples, particularly in the majority class. Unlike conventional approaches that employ a fixed-radius neighborhood in the input space, the proposed method computes the class probability using a kernel-based distance measure. The kernel-induced distance between two samples $a_i$ and $a_j$ is defined as $
d(a_i,a_j)
=
\sqrt{
2\big(1 - K(a_i,a_j)\big)
}.$

Let
\begin{equation}
\alpha_d = \max\!\left(r_{\max}^{(+)},\, r_{\max}^{(-)}\right)
\label{11}
\end{equation}
denote a data-dependent neighborhood radius, where $r_{\max}^{(+)}$ and $r_{\max}^{(-)}$ are the maximum kernel-induced radii of the positive and negative class samples, respectively.

For a given sample $a_i$, the local neighborhood is defined as $\mathcal{N}(a_i)
=
\{ a_j \;|\; d(a_i,a_j) \le \alpha_d \}.$ The class probability of $a_i$ is then computed as the ratio of samples belonging to the same class within this neighborhood:
\begin{equation}
p(a_i)
=
\frac{
\left|
\{ a_j \in \mathcal{N}(a_i) \;:\; b_j = b_k \}
\right|
}{
\left|
\mathcal{N}(a_i)
\right| + \varepsilon
},
\label{13}
\end{equation}
where $\varepsilon$ is a small positive constant introduced to avoid numerical instability.

A higher value of $p(a_i)$ indicates stronger local consistency of the sample with its class, whereas samples with low probability values are more likely to be noisy or located near class boundaries. Consequently, such samples are assigned lower influence during training, thereby improving the robustness of the proposed RoBell-RVFL model.

\subsubsection{PW-GB Membership Function}

By integrating the generalized bell (gbell) membership values with the local class probability, we propose a probability-weighted generalized bell (PW-GB) membership function. The PW-GB function is designed to suppress the influence of noisy and boundary samples in the majority class while preserving the full contribution of minority class samples.

The PW-GB membership function is defined as follows:
\begin{equation}
\xi(a_i) =
\begin{cases}
1, 
& \text{if } a_i = a_{i_{m}}, \\[6pt]
\xi_{\text{pw}}(a_i)\, p(a_i), 
& \text{if } a_i = a_i^{m}.
\end{cases}
\label{14}
\end{equation}

Here, $\xi_{\text{pw}}(a_i)$ represents the gbell membership value computed in the kernel-induced feature space, and $p(a_i)$ denotes the local class probability of the sample.
\subsection{Class Imbalance Weighting Scheme}
\label{www}\cite{{rezvani2022intuitionistic}}
The class-dependent weight $l_+ \text{ and }l_-/$ is defined as
\begin{equation}
l_- =1, \text{ if } a_i \text{ is in the class of negatives},
\label{eq:negative_weight}
\end{equation}
\begin{equation}
l_+ =  \frac{A_2}{A_1}, \quad \text{if } a_i  \text{ is in the class of positives}.
\label{eq:positive_weight}
\end{equation}

In the adopted weighting strategy, all samples belonging to the minority class are given a weight of one, whereas the samples from the majority class are down-weighted proportionally according to the ratio between the number of minority and majority instances.

\section{Proposed RoBell-RVFL}
\label{sec:bell_rvfl_cil}
This section presents a comprehensive formulation of the proposed RoBell-RVFL model. We begin by establishing a general mathematical framework designed to effectively manage class imbalance (CI). To mitigate the adverse effects of imbalanced data, the RoBell-RVFL model integrates a CI-aware weighting mechanism that adjusts sample contributions according to the imbalance ratio. Based on this formulation, the corresponding optimization problem of the proposed RoBell-RVFL model is expressed as follows:

\begin{equation}
\label{eq:opt_problem}
\begin{aligned}
min_{{V}} \; & 
\frac{1}{2}\|{V}\|_2^2
+ \frac{\gamma}{2}(l_{+})\left\|{L}_{+}^{\frac{1}{2}}{\beta}_{+}\right\|_2^2 \\
& + \frac{\gamma}{2}(l_{-})\left\|{L}_{-}^{\frac{1}{2}}{\beta}_{-}\right\|_2^2 \\
\text{s.t.} \quad &
{X}_{+}{V} = {Y}_{+} - {\beta}_{+}, \\
& {X}_{-}{V} = {Y}_{-} - {\beta}_{-}.
\end{aligned}
\end{equation}

Here, ${V}$ denotes the weight matrix that links both the input and hidden layers to the output layer. ${L}{-}$ and ${L}{+}$ are diagonal matrices whose diagonal elements correspond to the PW-GB membership values of the negative and positive class samples, respectively.

${L}_{+}= \text{diag}[ \xi(a_1^m), \xi(a_2^m), \cdots , \xi\!\left(a^{m}_{k+}\right) $, where $1 \leq i+ \leq A_1$ and ${L}_{-}= \text{diag}[ \xi(a_{1_{m}}), \xi(a_{2{_m}}), \cdots , \xi\!\left(a_{{i-}_m}\right)$, where $1 \leq i- \leq A_2$.

${\beta}_{-}$ (${\beta}_{+}$) denotes the error vector of the negative (positive) class samples. Here, $l_{-}$ ($l_{+}$) represents the weighting scheme of the negative (positive) class defined in Section \ref{www}. $\gamma \in \mathbb{R}^{+}$ is the regularization parameter used to penalize the error variables. ${X}_{-}$ (${X}_{+}$) denotes the nonlinear as well as linear projection of the negative (positive) class samples, defined as:
\begin{equation}
\label{eq:projections}
{X}_{-} = [{Z}_{-}\;{Y}_{-}], \quad
{X}_{+} = [{Z}_{+}\;{Y}_{\text{+}}].
\end{equation}

In this formulation, ${Z}{-}$ (${Z}{+}$) denotes the input data matrix corresponding to the negative (positive) class, while ${Y}{-}$ (${Y}{+}$) represents the hidden-layer output matrix for the negative (positive) class. These hidden representations are obtained by projecting ${Z}{-}$ (${Z}{+}$) through randomly initialized weights and biases, followed by the application of the nonlinear activation function $\phi$. The matrices ${Y}{-}$ and ${Y}{+}$ also serve as the target outputs for the negative and positive class samples, respectively.

Consequently, the incorporation of CI-aware weighting matrices ($l_{+}$ and $l_{-}$) enables the proposed model to regulate the influence of majority and minority class samples.

The corresponding Lagrangian formulation of \eqref{eq:opt_problem} is given by:
\begin{equation}
\label{eq:lagrangian}
\begin{aligned}
\mathcal{L} =\;
& \frac{1}{2}\|{V}\|_2^2
+ \frac{\gamma}{2}(l_{+})\left\|{L}_{+}^{\frac{1}{2}}{\beta}_{+}\right\|_2^2
+ \frac{\gamma}{2}(l_{-})\left\|{L}_{-}^{\frac{1}{2}}{\beta}_{-}\right\|_2^2 \\
& - {\alpha}_{+}^{\top}({X}_{+}{V} - {Y}_{+} + {\beta}_{+})
- {\alpha}_{-}^{\top}({X}_{-}{V} - {Y}_{-} + {\beta}_{-}),
\end{aligned}
\end{equation}
where ${\alpha}{-}$ and ${\alpha}{+}$ denote the Lagrange multipliers. 
Let
\begin{equation}
\label{eq:I_defs}
{F}_{+} = \gamma \times l_{+}, \quad {F}_{-} = \gamma \times l_{-}.
\end{equation}

Rewriting \eqref{eq:lagrangian}, we obtain:
\begin{equation}
\label{eq:lagrangian_rewritten}
\begin{aligned}
\mathcal{L} =\;
& \frac{1}{2}\|{V}\|_2^2
+  \frac{{F}_{+}}{2}\left\|{L}_{+}^{\frac{1}{2}}{\beta}_{+}\right\|_2^2
+ \frac{{F}_{-}}{2}\left\|{L}_{-}^{\frac{1}{2}}{\beta}_{-}\right\|_2^2 \\
& - 
\begin{bmatrix}
{\alpha}_{+}^{\top} & {\alpha}_{-}^{\top}
\end{bmatrix}
\left(
\begin{bmatrix}
{X}_{+} \\
{X}_{-}
\end{bmatrix}
{V}
-
\begin{bmatrix}
{Y}_{+} \\
{Y}_{-}
\end{bmatrix}
+
\begin{bmatrix}
{\beta}_{+} \\
{\beta}_{-}
\end{bmatrix}
\right).
\end{aligned}
\end{equation}

By applying the Karush--Kuhn--Tucker (KKT) conditions to \eqref{eq:lagrangian_rewritten}, we obtain:
\begin{align}
{V} - {X}_{+}^{\top}{\alpha}_{+} - {X}_{-}^{\top}{\alpha}_{-} &= 0, \label{eq:kkt_1} \\
{F}_{+}{L}_{+}{\beta}_{+} - {\alpha}_{+} &= 0, \label{eq:kkt_2} \\
{F}_{-}{L}_{-}{\beta}_{-} - {\alpha}_{-} &= 0, \label{eq:kkt_3} \\
{X}_{+}{V} - {Y}_{+} + {\beta}_{+} &= 0, \label{eq:kkt_4} \\
{X}_{-}{V} - {Y}_{-} + {\beta}_{-} &= 0. \label{eq:kkt_5}
\end{align}

Rewriting \eqref{eq:kkt_1}--\eqref{eq:kkt_3}, we have
\begin{align}
{I}{V}
&= {X}_{+}^{\top}{\alpha}_{+}
+ {X}_{-}^{\top}{\alpha}_{-}, \label{eq:beta_alpha} \\
{F}_{+}{L}_{+}{\beta}_{+}
&= {\alpha}_{+}, \label{eq:alpha_plus} \\
{F}_{-}{L}_{-}{\beta}_{-}
&= {\alpha}_{-}, \label{eq:alpha_minus}
\end{align}
where ${I}$ is an identity matrix of conformal dimension.
Using \eqref{eq:alpha_plus} and \eqref{eq:alpha_minus} in \eqref{eq:beta_alpha}, we obtain
\begin{align}
{I} {V}
&= {F}_{+}{X}_{+}^{\top}{L}_{+}{\beta}_{+}
+ {F}_{-}{X}_{-}^{\top}{L}_{-}{\beta}_{-}, \label{eq:beta_subst}
\end{align}
\begin{align}
{I}{V}
&= {F}_{+}{X}_{+}^{\top}{L}_{+}
({Y}_{+}-{X}_{+}{V})
+ {F}_{-}{X}_{-}^{\top}{L}_{-}
({Y}_{-}-{X}_{-}{V}), \label{eq:beta_expanded}
\end{align}
\begin{equation}
\label{eq:beta_grouped}
\resizebox{0.95\linewidth}{!}{$
({I}
+ {F}_{+}{X}_{+}^{\top}{L}_{+}{X}_{+}
+ {F}_{-}{X}_{-}^{\top}{L}_{-}{X}_{-})
{V}
= {F}_{+}{X}_{+}^{\top}{L}_{+}{Y}_{+}
+ {F}_{-}{X}_{-}^{\top}{L}_{-}{Y}_{-}. 
$}
\end{equation}

Multiplying $({F}_{+}^{-1}+{F}_{-}^{-1})$ to \eqref{eq:beta_grouped}, we have
\begin{align}
\Big(\frac{1}{{F}_{+}}+\frac{1}{{F}_{-}}\Big)
({I} 
+ {F}_{+}{X}_{+}^{\top}{L}_{+}{X}_{+}
+ {F}_{-}{X}_{-}^{\top}{L}_{-}{X}_{-})
{V} \notag \\
=
\Big(\frac{1}{{F}_{+}}+\frac{1}{{F}_{-}}\Big)
({F}_{+}{X}_{+}^{\top}{L}_{+}{Y}_{+}
+ {F}_{-}{X}_{-}^{\top}{L}_{-}{Y}_{-}). \label{eq:beta_mult}
\end{align}

Thus,
\begin{align}
{V}
&=
\Bigg(
\Big(\frac{1}{{F}_{+}}+\frac{1}{{F}_{-}}\Big)
({I} 
+ {X}_{+}^{\top}{L}_{+}{X}_{+}
+ {X}_{-}^{\top}{L}_{-}{X}_{-}
\Bigg)^{-1} \notag\\
&\quad \times
\Big(
{X}_{+}^{\top}{L}_{+}{Y}_{+}
+ {X}_{-}^{\top}{L}_{-}{Y}_{-}
\Big). \label{eq:beta_final_1}
\end{align}

Finally, we obtain the compact matrix form:
\begin{equation}
\label{eq:beta_compact}
\resizebox{0.99\linewidth}{!}{$
\begin{aligned}
{V}
&=
\Bigg(
\Big(\frac{1}{{F}_{+}}+\frac{1}{{F}_{-}}\Big){I}
+
[{X}_{+}^{\top}\;{X}_{-}^{\top}]
\begin{bmatrix}
\Big(1+\frac{{F}_{+}}{{F}_{-}}\Big){L}_{+} & {0} \\
{0} & \Big(1+\frac{{F}_{-}}{{F}_{+}}\Big){L}_{-}
\end{bmatrix}
\begin{bmatrix}
{X}_{+} \\
{X}_{-}
\end{bmatrix}
\Bigg)^{-1} \\[4pt]
&\quad \times
[{X}_{+}^{\top}\;{X}_{-}^{\top}]
\begin{bmatrix}
\Big(1+\frac{{F}_{+}}{{F}_{-}}\Big){L}_{+} & {0} \\
{0} & \Big(1+\frac{{F}_{-}}{{F}_{+}}\Big){L}_{-}
\end{bmatrix}
\begin{bmatrix}
{Y}_{+} \\
{Y}_{-}
\end{bmatrix}.
\end{aligned}
$}
\end{equation}

Using \eqref{eq:beta_compact}, we obtain the output layer weight matrix.

Section S.II, S.III, and S.V. of the Supplementary Material give the proposed methods' computational complexity, algorithm, and theoretical proof on the bound of the PW-GB membership function, respectively.

\begin{table*}[htbp]
\centering
\caption{Classification accuracies along with average accuracy and  average rankfor RoBell-RVFL model evaluated against baseline models on 26 UCI and KEEL datasets.}
\label{tab:multi_small}
\renewcommand{\arraystretch}{.6} 
\resizebox{\textwidth}{!}{%
\begin{tabular}{lcccccccc}
 \hline
\textbf{Dataset} $\downarrow$  \textbf{\textbar{} Model} $\rightarrow$ &
  \textbf{RVFL} \cite{PAO1994163} &
  \textbf{RVFLwoDL} \cite{HUANG2006489} &
  \textbf{NF-RVFL} \cite{10416391} &
  \textbf{C-RVFL} \cite{LIU2025112682} &
  \textbf{AC-RVFL} \cite{LIU2025112682} &
  \textbf{GB-RVFL} \cite{SAJID2025111142} &
  \textbf{GE-GB-RVFL} \cite{SAJID2025111142} &
  \textbf{RoBell-RVFL}$^{\dagger}$ \\
\hline

checkerboard\_Data &
  85.9423 &
  85.9808 &
  85.5769 &
  83.1731 &
  85.5769 &
  87.0192 &
  87.0192 &
  {87.5} \\
cleve &
  81.1111 &
  80 &
  82.2222 &
  73.3333 &
  81.1111 &
  75.5556 &
  82.2222 &
  {83.3333} \\
cmc &
  {68.552} &
  {68.552} &
  67.6471 &
  60.181 &
  57.6923 &
  68.0996 &
  {68.552} &
  66.9683 \\
congressional\_voting &
  61.0687 &
  61.0687 &
  51.9084 &
  61.0687 &
  {64.1221} &
  61.0687 &
  61.0687 &
  60.3053 \\
conn\_bench\_sonar\_mines\_rocks &
  74.6032 &
  71.4286 &
  74.6032 &
  65.0794 &
  66.6667 &
  74.6032 &
  68.254 &
  {82.5397} \\
crossplane130 &
  97.4359 &
  97.4359 &
  {100} &
  79.4872 &
  89.7436 &
  {100} &
  97.4359 &
  97.4359 \\
crossplane150 &
  81.1111 &
  81.1111 &
  88.8889 &
  88.8889 &
  73.3333 &
  86.6667 &
  73.3333 &
  {93.3333} \\
echocardiogram &
  85 &
  {87.5} &
  80 &
  85 &
  {87.5} &
  {87.5} &
  87 &
  82.5 \\
ecoli0137vs26 &
  85.7447 &
  86.8085 &
  {94.6809} &
  82.9787 &
  87.234 &
  87.234 &
  88.2979 &
  {94.6809} \\
ecoli2 &
  82.0792 &
  81.0891 &
  88.1188 &
  {89.1089} &
  {89.1089} &
  87.1287 &
  87.1287 &
  88.1188 \\
fertility &
  {90} &
  80 &
  73.3333 &
  {90} &
  {90} &
  86.6667 &
  {90} &
  86.6667 \\
haberman &
  76.087 &
  76.087 &
  77.1739 &
  {82.6087} &
  {82.6087} &
  78.2609 &
  77.1739 &
  76.087 \\
haberman\_survival &
  76.087 &
  78.2609 &
  71.7391 &
  {82.6087} &
  {82.6087} &
  78.2609 &
  77.1739 &
  76.087 \\
heart\_hungarian &
  72.7753 &
  72.6517 &
  {79.7753} &
  75.2809 &
  76.4045 &
  74.1573 &
  74.1573 &
  75.2809 \\
heart-stat &
  81.8889 &
  81.8889 &
  {87.6543} &
  77.7778 &
  82.716 &
  82.7161 &
  81.4815 &
  {87.6543} \\
iono &
  88.6792 &
  80.566 &
  83.9623 &
  86.7925 &
  88.6792 &
  83.9623 &
  84.9057 &
  {90.566} \\
ionosphere &
  82.7925 &
  82.4528 &
  {87.7358} &
  85.8491 &
  83.9623 &
  83.9623 &
  83.0189 &
  85.8491 \\
led7digit-0-2-4-5-6-7-8-9\_vs\_1 &
  94.7368 &
  94.7368 &
  92.4812 &
  93.2331 &
  93.2331 &
  93.985 &
  94.7368 &
  {96.2406} \\
molec\_biol\_promoter &
  71 &
  71.875 &
  62.5 &
  62.5 &
  71.875 &
  71.875 &
  {75} &
  65.625 \\
monk1 &
  42.515 &
  52.0958 &
  44.3114 &
  43.1138 &
  44.3114 &
  44.3114 &
  47.9042 &
  {53.8922} \\
monk3 &
  43.1138 &
  43.1138 &
  {76.0884} &
  43.7126 &
  45.509 &
  44.9102 &
  45.509 &
  45.509 \\
monks\_3 &
  90.4072 &
  90.4072 &
  90.6347 &
  74.8503 &
  85.0299 &
  85.6287 &
  91.018 &
  {97.006} \\
oocytes\_merluccius\_nucleus\_4d &
  80.7134 &
  {82.7362} &
  75.1792 &
  67.4267 &
  77.1987 &
  80.7818 &
  78.5016 &
  78.1759 \\
vehicle1 &
  81.6772 &
  81.8583 &
  {83.8583} &
  77.5591 &
  77.9528 &
  81.1024 &
  79.9213 &
  {83.8583} \\
vehicle2 &
  90.8504 &
  90.0315 &
  {97.6378} &
  88.189 &
  86.6142 &
  91.7323 &
  92.5197 &
  {97.6378} \\
yeast-0-2-5-7-9\_vs\_3-6-8 &
  95.351 &
  89.7086 &
  95.6954 &
  91.0596 &
  91.3907 &
  {98.0133} &
  97.351 &
  96.3576 \\
  \hline
{\textbf{Average accuracy}} &
79.2817 &	78.8248	&80.5156&	76.5716&	78.5455&	79.8155&	79.6417&	\textbf{81.8927} \\

  \hline
{\textbf{Average rank}} &
5.0385&	5.0769&	4.2692&	5.7308&	4.5962&	4.0577&	4&	\textbf{3.2308}\\
\hline
  \multicolumn{8}{l}{$^{\dagger}$ represents the proposed models. The boldface indicates the best in terms of accuracy.}
\end{tabular}%
}
\end{table*}



\subsection{Evaluation Dataset}  
Table~\ref{tab:multi_small} summarizes the classification accuracies and average ranks of all compared methods on 26 benchmark datasets. Detailed results, including standard deviations and optimal hyperparameter settings for each dataset, are provided in the Supplementary Material (Table IV). Among all evaluated models, the proposed {RoBell-RVFL} achieves the highest average classification accuracy of {81.8927\%}, outperforming the second-best method, NF-RVFL, which attains an average accuracy of 80.5156\%. These results demonstrate the consistent superiority of RoBell-RVFL over both baseline RVFL models and recent state-of-the-art variants.

In comparison, conventional RVFL-based methods, namely RVFL and RVFLwoDL, obtain lower average accuracies of 79.2817\% and 78.8248\%, respectively. Complex-valued extensions, including CRVFL and ACRVFL, further lag behind with average accuracies of 76.5716\% and 78.5455\%. Similarly, granular ball-based approaches, GE-GB-RVFL and GB-RVFL, achieve average accuracies of 79.6417\% and 79.8155\%, respectively. Overall, these results highlight the superior generalization capability and stable performance gains of RoBell-RVFL across diverse datasets, confirming its effectiveness in comparison with both classical and modern RVFL-based models.

\noindent
\textbf{Statistical rank:} Since average accuracy may be biased by outlier datasets, a ranking-based evaluation is adopted following~\cite{demvsar2006statistical}, where models are ranked independently on each dataset and lower ranks indicate better performance. Let \( \delta \) denote the number of models evaluated over \( \omega \) datasets, and \( R_r^s \) represent the rank of the \( r^{{th}} \) model on the \( s^{{th}} \) dataset. The average rank is computed as \( R_r = \frac{1}{\omega} \sum_{s=1}^{\omega} R_r^s \). Using this criterion, the proposed {RoBell-RVFL} achieves the lowest average rank ({3.2308}), outperforming all competing methods. In contrast, GE-GB-RVFL (\(4.0000\)) and GB-RVFL (\(4.0577\)) rank second and third, respectively, while NF-RVFL (\(4.2692\)), CRVFL (\(5.7308\)), and ACRVFL (\(4.5962\)) exhibit higher rank values, confirming the superior robustness and generalization capability of RoBell-RVFL across diverse datasets.

\noindent
\textbf{Friedman test:} The Friedman test~\cite{friedman1940comparison} is employed to statistically examine whether significant performance differences exist among the compared models. Under the null hypothesis, all models are assumed to have equal average ranks, indicating equivalent performance. The Friedman statistic follows a chi-squared distribution, denoted as \(\chi^2_F\), with \((\delta - 1)\) degrees of freedom and is computed as$\chi^2_F = \frac{12\omega}{\delta(\delta + 1)} 
\left( \sum_{i=1}^{\delta} R_i^2 - \frac{\delta(\delta + 1)^2}{4} \right),$ where \(\delta\) denotes the number of models, \(\omega\) represents the number of datasets, and \(R_i\) is the average rank of the \(i^{\text{th}}\) model. The corresponding F-statistic is given by $F_F = \frac{(\omega - 1)\chi^2_F}{\omega(\delta - 1) - \chi^2_F},$ which follows an \(F\)-distribution with \((\delta - 1)\) and \((\omega - 1)(\delta - 1)\) degrees of freedom. In our experimental setting, eight models are evaluated across 26 datasets, yielding \(\chi^2_F = 18.4495\) and \(F_F = 2.8202\). At a 5\% significance level, the critical value of the \(F\)-distribution with \((7,175)\) degrees of freedom is 2.0622. Since \(F_F > 2.0622\), the null hypothesis is rejected, confirming the existence of statistically significant performance differences among the compared models.  

In Supplementary Material Section S.VI. and Section S.VIII., discusses Win–tie–loss (W–T–L) sign test and sensitivity analysis, respectively.

\section{Conclusion and Future Work}
\label{Sec:Conclusion}

This paper introduced {RoBell-RVFL}, a probability-weighted generalized bell random vector functional link network designed to address the fundamental challenges of class imbalance, label noise, and outliers in real-world data. Unlike conventional RVFL-based models that assign uniform importance to all training samples, RoBell-RVFL employs a principled sample-level adaptive weighting mechanism by jointly integrating kernel-induced generalized bell (gbell) membership functions with local class probability estimation. This strategy enables effective suppression of noisy and boundary samples within the majority class while fully preserving minority class information, resulting in more reliable and unbiased decision boundaries. Notably, class imbalance is handled implicitly through class-dependent regularization, avoiding heuristic or ratio-based weighting schemes.

Extensive experiments on 26 benchmark datasets from the UCI and KEEL repositories demonstrate that RoBell-RVFL consistently outperforms state-of-the-art RVFL variants, neuro-fuzzy models, complex-valued RVFLs, and recent granular and graph-based approaches in terms of classification accuracy, average rank, robustness to label noise, and statistical significance. These improvements are further supported by Friedman and win–tie–loss statistical tests, confirming that the observed performance gains are consistent and statistically meaningful. Overall, the results highlight that effective learning from imbalanced data requires simultaneous consideration of sample quality, class dominance, and noise suppression capabilities that are jointly realized by the performance of the proposed RoBell-RVFL framework.

\bibliographystyle{IEEEtran}
\bibliography{ref}

\clearpage
\section*{Supplementary Material}

\section*{S.I. Related Works}\label{Sec:Related_work}
In this section, we discuss on the framework of RVFL mathematically.
\subsection{RVFL}
RVFL neural network consists of three layers: the input layer, hidden layer, and output layer. The weights between the input and hidden layers, as well as the hidden layer biases, are randomly initialized and remain fixed throughout the training process.
 
The output layer receives features from both the input layer (via direct connections) and the hidden layer. The output weights are typically determined using techniques like the Moore-Penrose pseudoinverse or the least squares method. 
Figure \ref{fig:rvfl_architecture} depicts the architectural layout of the RVFL model.

\renewcommand{\thefigure}{S.1}
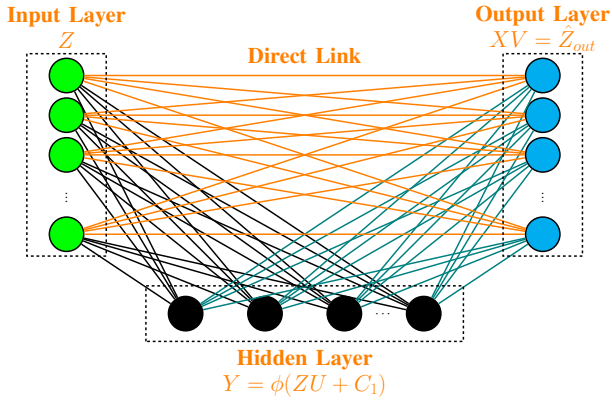
\begin{figure}[htbp]
    \centering
    
\scalebox{0.35}{%

\begin{tikzpicture}[x=3cm, y=1.5cm, >=stealth]

\foreach \i in {1,...,3} {
    \node[circle, draw, fill=green!100, minimum size=1.3cm, ultra thick] (I-\i) at (0, -\i) {};
    \node[anchor=east, font=\large] at (I-\i.west) {$ $};
}
\node[align=center, font=\Large] at (0, -4) {$\vdots$};
\node[circle, draw, fill=green!100, minimum size=1.3cm, ultra thick] (I-4) at (0, -5) {};
\node[anchor=east, font=\large] at (I-4.west) {$ $};

\foreach \i in {1,2,3} {
    \node[circle, draw, fill=gray!200, minimum size=1.3cm, ultra thick] (H-\i) at (.5+\i, -7) {};
}
\node[align=center, font=\Large] at (4, -7) {$\cdots$};
\node[circle, draw, fill=gray!200, minimum size=1.3cm, ultra thick] (H-4) at (4.5, -7) {};

\foreach \i in {1,...,3} {
    \node[circle, draw, fill=cyan!100, minimum size=1.3cm, ultra thick] (O-\i) at (6, -\i) {};
    \node[anchor=west, font=\large] at (O-\i.east) {$ $};
}
\node[align=center, font=\Large] at (6, -4) {$\vdots$};
\node[circle, draw, fill=cyan!100, minimum size=1.3cm, ultra thick] (O-4) at (6, -5) {};
\node[anchor=west, font=\large] at (O-4.east) {$ $};

\foreach \i in {1,...,4} {
    \foreach \j in {1,...,4} {
        \draw[black, ultra thick] (I-\i) -- (H-\j);
    }
}

\foreach \i in {1,...,4} {
    \foreach \j in {1,...,4} {
        \draw[teal, ultra thick] (H-\i) -- (O-\j);
    }
}

\foreach \i in {1,...,4} {
    \foreach \j in {1,...,4} {
        \draw[orange!100, ultra thick] (I-\i) -- (O-\j);
    }
}

\node[orange!100, align=center, font=\Huge\bfseries] at (-0, 0.2) {\textbf{Input Layer} \\ ${Z}$};
\node[orange!100, align=center, font=\Huge\bfseries] at (6, 0.2) {\textbf{Output Layer} \\ $X{V} = {\hat{Z}_{out}}$};
\node[orange!100, align=center, font=\Huge\bfseries] at (3, -8.5) {\textbf{Hidden Layer} \\ $Y = \phi({Z}{U} + {C}_1)$};

\node[orange!100, align=center, rotate=0, font=\Huge\bfseries] at (3, -.5) {\textbf{Direct Link}};

\draw[dashed, ultra thick] (-0.5, -0.45) rectangle (0.5, -5.55);
\draw[dashed, ultra thick] (1, -7.7) rectangle (5, -6.3);
\draw[dashed, ultra thick] (5.5, -0.45) rectangle (6.5, -5.55);

\end{tikzpicture}

}
    \caption{Architecture of the RVFL neural network.}
    \label{fig:rvfl_architecture}
\end{figure}

The hidden layer matrix, represented by \({Y}\) and its mathematical representation is:
\begin{equation}
{Y} = \phi({Z} U + {C}_1)  \in \mathbb{R}^{A \times D}\text{,}
\label{eq:1}
\end{equation}
where \(U \in \mathbb{R}^{B \times D}\) is a matrix of weights, initialized randomly from a uniform distribution in the range \([-1, 1]\), \({C}_1 \in \mathbb{R}^{A \times D}\) is the bias matrix and $\phi$ is an activation function.

The original input features are concatenated with the hidden layer features to form the matrix \(X\), expressed as:
\begin{equation}
X = \begin{bmatrix} {Z} \mid Y \end{bmatrix} \in \mathbb{R}^{A \times (B+D)}.
\label{eq:concatenated_features}
\end{equation}
The predicted output ${{{\hat{Z}_{out}}}}$ is given by:
\begin{equation}
{\hat{Z}_{out}} = X V,
\end{equation}
where \(V \in \mathbb{R}^{(B+D) \times E}\) is the weight matrix connecting the concatenated feature matrix to the output layer.
\section*{S.II. Computational Complexity}
Let $A_1$ and $A_2$ be the numbers of majority and minority
class samples, respectively, such that $A=A_1+A_2$.

The proposed RoBell-RVFL training consists of PW-GB membership
computation and closed-form RVFL learning.

The PW-GB membership computation employs a Gaussian kernel and requires
pairwise similarity evaluation among samples. The kernel matrix
construction for majority and minority classes has a time complexity of $\mathcal{O}(A_1^2 B) + \mathcal{O}(A_2^2 B)
= \mathcal{O}(A^2 B).$ The output weight matrix $\mathbf{V}$ is obtained by solving
a linear system of size $(B+D)$, which incurs a computational cost of $\mathcal{O}\!\left(A(B+D)^2 + (B+D)^3\right).$ Therefore, the overall training time complexity of the proposed
RoBell-RVFL model is $\mathcal{O}
A^2 B
+
A B D
+
A(B+D)^2
+
(B+D)^3.$

\section*{S.III. Algorithm of the proposed RoBell-RVFL}

\begin{algorithm}[htbp]
\caption{RoBell-RVFL}
\label{alg:RoBell-RVFL}
\begin{algorithmic}[1]
\State \textbf{Input}: Training data 
$\mathcal{D}=\{(a_i,b_i)\}_{i=1}^{A}$, 
$a_i\in\mathbb{R}^{B}$, $b_i\in\{+1,-1\}$;  
number of hidden nodes $D$;  
kernel width $\xi$;  
gbell parameter $\rho$;  
regularization parameter $\gamma$.
\State \textbf{Output}: Weight matrix $V$.

\State Partition data into majority and minority class sets
$T_1$ and $T_2$ with sizes $A_1$ and $A_2$.

\State Compute Gaussian kernel matrix
$K(a_i,a_j)$.

\State Compute kernel-induced radii
$r_i^{(+)}$ and $r_i^{(-)}$ using Eqs. 1-2 of the main manuscript, and obtain
$r_{\max}^{(+)}$, $r_{\max}^{(-)}$ using Eq. 3 of the main manuscript.

\State Compute gbell memberships
$\xi_{\mathrm{pw}}^{(+)}(a_i)$ and $\xi_{\mathrm{pw}}^{(-)}(a_i)$
using Eqs. 4-5 of the main manuscript.

\State Compute neighborhood radius $\alpha_d$
using Eq. 6 of the main manuscript.

\State Compute class probability $p(a_i)$ using Eq. 7 of the main manuscript.

\State Compute PW-GB membership values $\xi(a_i)$ using Eq. 8 of the main manuscript and form diagonal matrices $L_+$ and $L_-$.

\State Assign class imbalance weights $l_+$ and $l_-$ using
Eqs. 14 of the main manuscript, and compute $F_+=\gamma l_+$, $F_-=\gamma l_-$.


\State Compute hidden layer output
$Y=\phi(ZU+C_1)$ using Eq.~\eqref{eq:1} and construct
$X=[Z\mid Y]$.

\State Compute output weight matrix $V$ using the closed-form
solution in Eq. 29 of the main manuscript.

\Return $V$.
\end{algorithmic}
\end{algorithm}


\section*{S.IV. Experimental Setup, Compared Models and Datasets}
\label{Sec:experiment}

\textbf{Experimental Setup:} All experiments are conducted on a workstation equipped with an Intel(R) Xeon(R) Gold 6226R processor operating at 2.90 GHz and 128 GB of RAM, running on the Windows 11 operating system. The proposed and comparative models are implemented in Python 3.11. For performance evaluation, each dataset is randomly partitioned into training and testing sets using a 70:30 split. A five-fold cross-validation scheme, coupled with a grid search strategy, is employed to determine the optimal hyperparameter settings. Specifically, the search ranges are defined as $\gamma = {10^{-5}, 10^{-4}, \ldots, 10^{5}}$, $\xi = {2^{-5}, 2^{-4}, \ldots, 2^{5}}$, and $\rho = {\tfrac{1}{2}, \tfrac{5}{8}, \tfrac{3}{4}, \tfrac{7}{8}, 1}$. The number of hidden nodes is varied within the range $3:20:203$. Consistent with \cite{sajid2024neuro}, the NF-RVFL model employs a neuro-fuzzy layer with the number of fuzzy rules selected from $J = 5:5:50$, where the cluster centers are determined using the $k$-means clustering algorithm. For the GB-RVFL and GE-GB-RVFL models, the parameter configurations follow those reported in \cite{SAJID2025111142}. Additionally, for all applicable models, ten different activation functions (indexed from 1 to 10) are considered during hyperparameter tuning. A comprehensive summary of these configurations is provided in Table~S.1 of the supplementary material.
\\
\textbf{Compared Models:}
     The selected baseline models include the standard RVFL~\cite{PAO1994163}, neuro-fuzzy RVFL (NF-RVFL)~\cite{sajid2024neuro}, RVFL without direct links (RVFLwoDL), also known as the extreme learning machine (ELM)~\cite{HUANG2006489}, complex-valued RVFL (CRVFL)~\cite{LIU2025112682}, augmented complex-valued RVFL (ACRVFL)~\cite{LIU2025112682}, granular ball RVFL (GB-RVFL)~\cite{SAJID2025111142}, and graph-embedded GB-RVFL (GE-GB-RVFL)~\cite{SAJID2025111142}.

\textbf{Datasets:} To evaluate the performance of our proposed model, we conduct a comprehensive comparison against a diverse set of randomized neural network (RdNN) based architectures.


\subsection*{ Performance Metrics}

To rigorously evaluate the effectiveness of the proposed PW-RVFL-CIL framework, its performance is assessed using several widely adopted classification metrics, namely Accuracy, Sensitivity, Specificity, Precision, F-measure, and G-mean. These metrics collectively provide a comprehensive and balanced evaluation of predictive performance, particularly in the presence of class imbalance and label noise.

The mathematical definitions of the employed performance measures are given as follows:

\begin{equation}
\text{Accuracy} =
\frac{\text{True}_{+} + \text{True}_{-}}
{\text{True}_{+} + \text{False}_{+} + \text{True}_{-} + \text{False}_{-}},
\end{equation}

\begin{equation}
\text{Sensitivity} =
\frac{\text{True}_{+}}
{\text{True}_{+} + \text{False}_{-}},
\end{equation}

\begin{equation}
\text{Specificity} =
\frac{\text{True}_{-}}
{\text{True}_{-} + \text{False}_{+}},
\end{equation}

\begin{equation}
\text{Precision} =
\frac{\text{True}_{+}}
{\text{True}_{+} + \text{False}_{+}},
\end{equation}

\begin{equation}
\text{F-measure} =
\frac{2 \times \text{Precision} \times \text{Sensitivity}}
{\text{Precision} + \text{Sensitivity}},
\end{equation}

\begin{equation}
\text{G-mean} =
\sqrt{\text{Sensitivity} \times \text{Specificity}}.
\end{equation}

These metrics characterize complementary aspects of classification performance. Here, $\text{True}_{+}$ and $\text{True}_{-}$ denote the number of correctly classified positive and negative samples, respectively, while $\text{False}_{+}$ and $\text{False}_{-}$ represent false positive and false negative outcomes. The inclusion of F-measure and G-mean ensures a fair and informative evaluation by jointly considering both class-wise performance and balance between sensitivity and specificity, which is particularly important for imbalanced learning scenarios.

\section*{S.V. Theoretical proof on the bound of the PW-GB membership function}
\textbf{Theorem 1:}
For all training samples $a_i$, the  PW-GB membership function satisfies
\begin{equation}
0 \le \xi(a_i) \le 1, \quad k = 1,2,\ldots,n.
\end{equation}

\textbf{Proof:}
The proof is divided into two cases.

\textit{Case 1: Minority class samples.}  
For any minority class sample $a_{i_{m}}$, the PW-GB membership is defined as
\[
\xi(a_{i_{{m}}}) = 1.
\]
Hence, the condition $0 \le \xi(a_{i_{{m}}}) \le 1$ holds trivially.

\textit{Case 2: Majority class samples.}  
For any majority class sample $a_i^{m}$, the PW-GB membership is defined as
\[
\xi(a_i^{{m}}) = \xi_{\text{pw}}(a_i)\, p(a_i).
\]

From the gbell membership definition used in the proposed method,
\[
\xi_{\text{pw}}(a_i)
=
\frac{1}{1+\left(\frac{r_k}{c}\right)^{2d}},
\]
where $r_i \ge 0$ denotes the kernel-induced radius of sample $a_i$, and $c>0$, $d>0$ are positive parameters. Since $\left(\frac{r_i}{c}\right)^{2d} \ge 0$, it follows that
\[
0 < \xi_{\text{pw}}(a_i) \le 1.
\]

Furthermore, the class probability $p(a_i)$ is computed as the ratio of the number of samples belonging to the same class within a local neighborhood to the total number of samples in that neighborhood. Therefore,
\[
0 \le p(a_i) \le 1.
\]

Since the product of two values lying in $[0,1]$ also lies in $[0,1]$, we have
\[
0 \le \xi(a_i^{m}) \le 1.
\]

Combining both cases, the proposed PW-GB membership function satisfies
\[
0 \le \xi(a_i) \le 1, \quad \forall i,
\]
which completes the proof.

\renewcommand{\thetable}{S.I}
\begin{table*}[htbp]
    \caption{Pairwise win-tie-loss test of proposed models and baseline models on MCD category UCI datasets.}
    \label{win tie loss sign test for linear}
    \resizebox{1.0\textwidth}{!}{
\begin{tabular}{lccccccccccccc}
\hline
\multicolumn{1}{c}{} &
\multicolumn{1}{c}{\textbf{RVFLwoDL} \cite{HUANG2006489}  } &
\multicolumn{1}{c}{ \textbf{RVFL} \cite{PAO1994163} } &
\multicolumn{1}{c}{\textbf{NF-RVFL} \cite{sajid2024neuro} } &
\multicolumn{1}{c}{ CRVFL \cite{LIU2025112682}}&
\multicolumn{1}{c}{  ACRVFL \cite{LIU2025112682}}&

\multicolumn{1}{c}{\textbf{GB-RVFL} \cite{SAJID2025111142}}  &
\multicolumn{1}{c}{\textbf{GE-GB-RVFL} \cite{SAJID2025111142}}  &



\\

\hline

 \textbf{RVFL} \cite{PAO1994163} & [$9,9,8$] 	 &  &  \\

\textbf{NF-RVFL} \cite{sajid2024neuro} & [$16,0,10$] & [$15,1,10$] 	&  &  &  &  &  \\

\textbf{C-RVFL} \cite{LIU2025112682} & [$10,1,15$] & [$8,3,15$] & [$8,2,16$] & [$9,2,17$]	 &  &  &  \\

\textbf{AC-RVFL} \cite{LIU2025112682} & [$13,2,11$] & [$12,3,11$] & [$10,2,14$] & [$17,5,4$]   \\

 \textbf{GB-RVFL} \cite{SAJID2025111142} & [$15,4,7$] & [$17,2,7$] & [$11,4,11$] & [$17,1,8$] & [$12,5,9$]

\\

 \textbf{GE-GB-RVFL} \cite{SAJID2025111142} & [$14,4,8$] & [$15,5,6$] & [$13,2,11$] & [$17,2,7$] & [$14,3,9$] & [$11,4,11$]

\\

{\textbf{RoBell-RVFL}$^{\dagger}$} &\textbf{{[$18,2,6$]}} & [$17,3,6$] & [$15,5,6$] & \textbf{{[$18,2,6$]}} & [$17,1,8$] & [$16,1,9$] & [$15,2,9$]	 \\

\hline
 \multicolumn{7}{l}{$^{\dagger}$ represents the proposed models.}
\end{tabular}}
\end{table*}

\section*{S.VI. Win–tie–loss (W–T–L) sign test:} To further analyze pairwise performance differences, we employ the win–tie–loss (W–T–L) sign test~\cite{demvsar2006statistical}. Table~\ref{win tie loss sign test for linear} summarizes the comparative results of the proposed RoBell-RVFL against baseline models on the UCI and KEEL datasets. In this table, each entry \([x, y, z]\) denotes the number of datasets on which the model in the corresponding row wins, ties, and loses, respectively, against the model listed in the column.

Under the null hypothesis, two models are assumed to perform equally well, implying that each model is expected to win on approximately \(\omega/2\) out of \(\omega\) datasets. A statistically significant difference is established when a model achieves at least \(\omega/2 + 1.96\sqrt{\omega/2}\) wins. In the presence of ties, wins are evenly distributed; if the number of ties is odd, one tie is discarded prior to distribution. For \(\omega = 26\), the resulting threshold is 18 wins.

Based on this criterion, RoBell-RVFL attains 18 wins against RVFLwoDL and CRVFL, demonstrating statistically significant superiority over these models. Furthermore, RoBell-RVFL records 17 wins against RVFL and ACRVFL, 16 wins against GB-RVFL, and 15 wins against NF-RVFL and GE-GB-RVFL. Overall, the W–T–L analysis confirms that RoBell-RVFL consistently outperforms the majority of baseline methods, reinforcing its robustness and competitive advantage across diverse benchmark datasets.

\renewcommand{\thetable}{S.II}
\begin{table}[htbp]
\caption{Classification performance of models on different levels of label noise}
\label{noise_table}
\resizebox{0.5\textwidth}{!}{%
\begin{tabular}{ccccc}
\hline
\multicolumn{1}{l}{}                & Noise   & RoBell-RVFL\cite{SAJID2025111142}      & RVFL \cite{PAO1994163}           & GB-RVFL \cite{SAJID2025111142}          \\
\hline
\multirow{5}{*}{heart-stat}         & 5\%     & \textbf{85.1852} & 85.1235          & 54.321           \\
                                    & 10\%    & \textbf{90.1235} & 82.7161          & 86.4198          \\
                                    & 20\%    & 81.4815          & \textbf{83.9506} & 82.7161          \\
                                    & 30\%    & 81.4815          & 56.7901          & \textbf{82.7161} \\
                                    & 40\%    & 49.3827          & 50.6173          & \textbf{77.7778} \\
                                    \hline
\multicolumn{2}{c}{\textbf{Average accuracy}} & \textbf{77.5309} & 71.8395          & 76.7902          \\
\hline
\multirow{5}{*}{heart\_hungarian}   & 5\%     & \textbf{78.6517} & 74.1573          & 75.2809          \\
                                    & 10\%    & \textbf{77.5281} & 71.9101          & 74.1573          \\
                                    & 20\%    & \textbf{82.0225} & 76.4045          & 80.8989          \\
                                    & 30\%    & 66.2921          & 61.7978          & \textbf{67.4157} \\
                                    & 40\%    & \textbf{64.0449} & \textbf{64.0449} & 55.0562          \\
                                    \hline
\multicolumn{2}{c}{\textbf{Average accuracy}} & \textbf{73.7079} & 69.6629          & 70.5618          \\
\hline
\multirow{5}{*}{led7digit-0-2-4-5-6-7-8-9\_vs\_1} & 5\% & \textbf{95.4887} & 91.7368 & 83.4587 \\
                                    & 10\%    & 93.985           & \textbf{94.7368} & \textbf{94.7368} \\
                                    & 20\%    & \textbf{93.2331} & 83.985           & 73.6842          \\
                                    & 30\%    & \textbf{87.9699} & 84.2105          & 61.6541          \\
                                    & 40\%    & \textbf{88.7218} & 70.4511          & 43.609           \\
                                    \hline
\multicolumn{2}{c}{\textbf{Average accuracy}} & \textbf{91.8797} & 85.024           & 71.4286          \\
\hline
\multirow{5}{*}{monk1}              & 5\%     & \textbf{53.8922} & 46.7066          & 43.7126          \\
                                    & 10\%    & 48.503           & 48.503           & \textbf{50.2994} \\
                                    & 20\%    & 48.503           & 47.9042          & \textbf{51.497}  \\
                                    & 30\%    & 45.509           & \textbf{52.0958} & 47.3054          \\
                                    & 40\%    & \textbf{52.0958} & 48.8982          & 49.7006          \\
                                    \hline
\multicolumn{2}{c}{\textbf{Average accuracy}} & \textbf{49.7006} & 48.8216          & 48.503           \\
\hline
\multirow{5}{*}{vehicle2}           & 5\%     & 94.8819          & \textbf{97.2441} & 92.9134          \\
                                    & 10\%    & 95.2756          & 95.2756          & \textbf{96.063}  \\
                                    & 20\%    & 90.5512          & \textbf{93.7008} & 93.3071          \\
                                    & 30\%    & 85.4331          & \textbf{94.0945} & 92.5197          \\
                                    & 40\%    & \textbf{76.378}  & 60.7087          & 60.2362          \\
                                    \hline
\multicolumn{2}{c}{\textbf{Average accuracy}} & \textbf{88.504}  & 88.2047          & 87.0079          \\
\hline
\multicolumn{2}{c}{\textbf{Overall average accuracy}}   & \textbf{78.4137} & 76.761  & 76.4659\\
\hline
\end{tabular}
}

\end{table}

\section*{S.VII. Evaluation of RoBell-RVFL under Label Noise}
To rigorously assess the robustness of the proposed RoBell-RVFL model, five datasets with diverse structural and statistical characteristics are randomly selected. As reported in Table~\ref{noise_table} and illustrated in Figure~\ref{noise_plot}, RoBell-RVFL consistently outperforms the baseline RVFL across nearly all datasets and noise levels.

Across all evaluated noise conditions, RoBell-RVFL demonstrates strong resilience to label corruption, maintaining stable performance even as the noise ratio increases. Notably, RoBell-RVFL achieves the highest overall average accuracy of \textbf{78.4137\%}, compared to 76.7610\% for the baseline RVFL and 76.4659\% for GB-RVFL, indicating a clear improvement in predictive reliability. This advantage is further corroborated by the dataset-wise average accuracies in Table~\ref{noise_table}, where the proposed model consistently surpasses competing methods across all evaluated datasets.

Overall, these results confirm that RoBell-RVFL offers superior noise tolerance, enhanced generalization capability, and more stable performance under varying levels of label noise, underscoring its effectiveness in noisy and challenging learning environments.

\renewcommand{\thefigure}{S.2}
\begin{figure*}[htp]
\begin{minipage}{.300\linewidth}
\centering
\subfloat[heart\_hungarian\label{3d1}]{\includegraphics[scale=0.20]{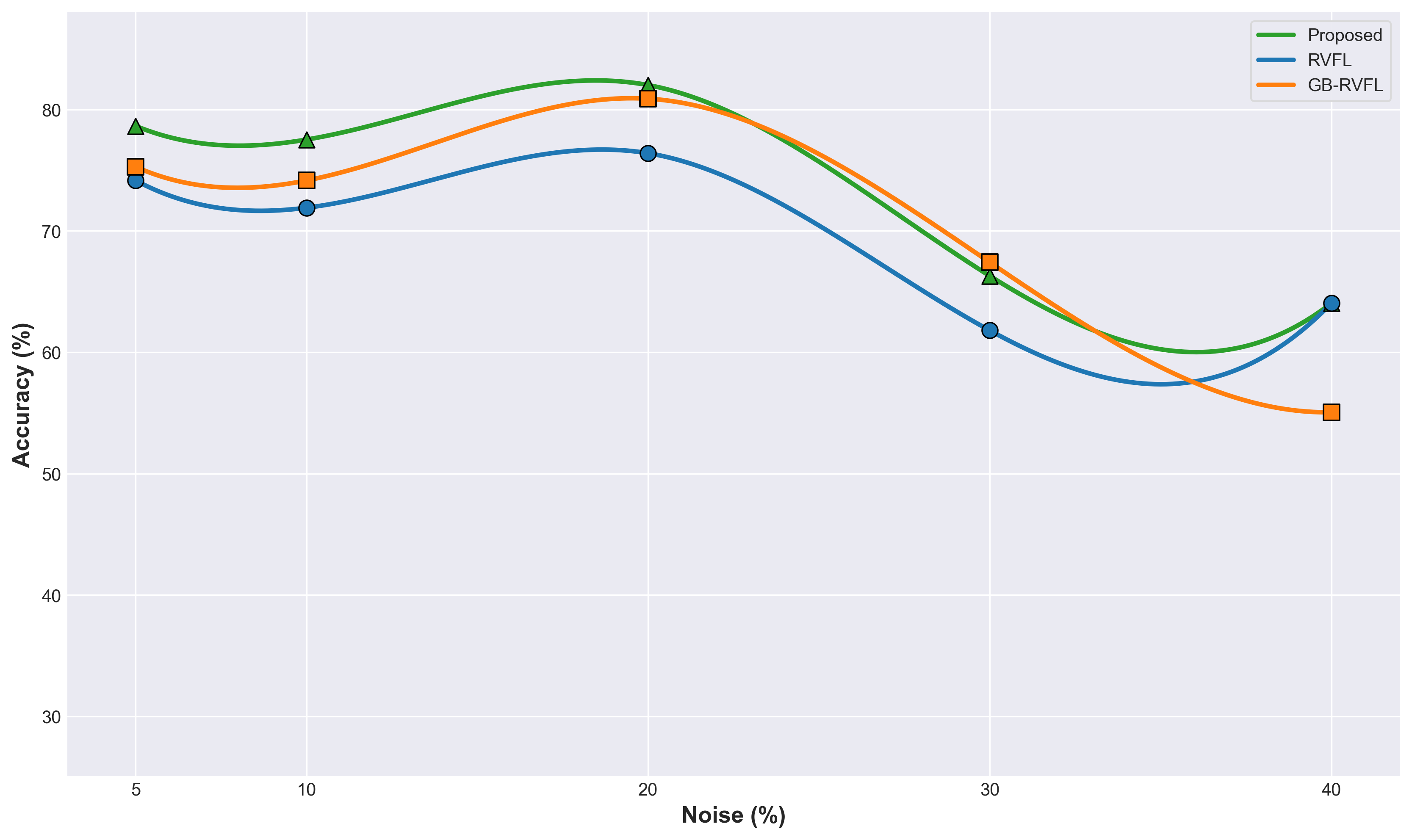}}
\end{minipage}
\begin{minipage}{.300\linewidth}
\centering
\subfloat[vehicle2\label{3d2}]{\includegraphics[scale=0.20]{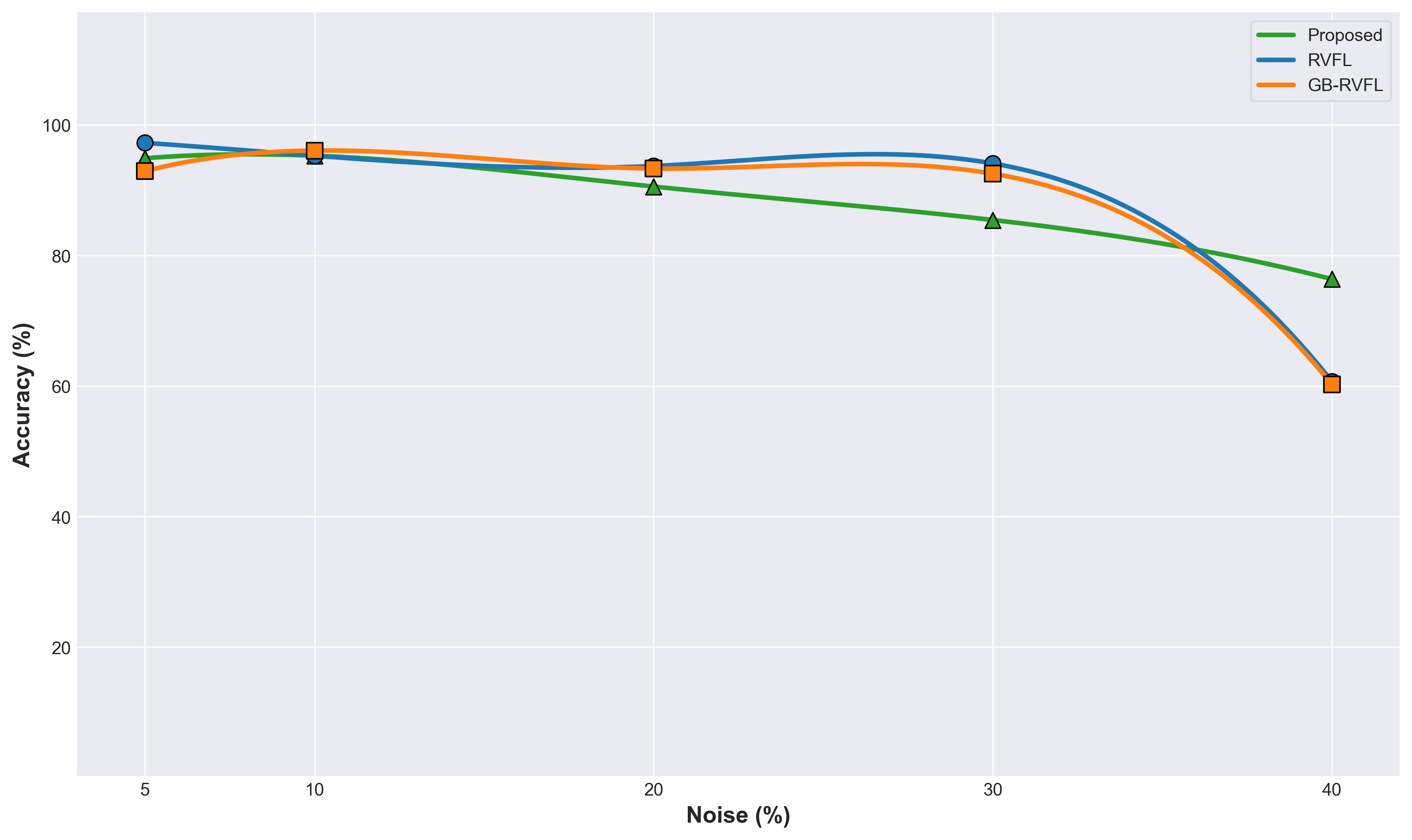}}
\end{minipage}
\begin{minipage}{.300\linewidth}
\centering
\subfloat[led7digit-0-2-4-5-6-7-8-9\_vs\_1\_3\label{3d3}]{\includegraphics[scale=0.20]{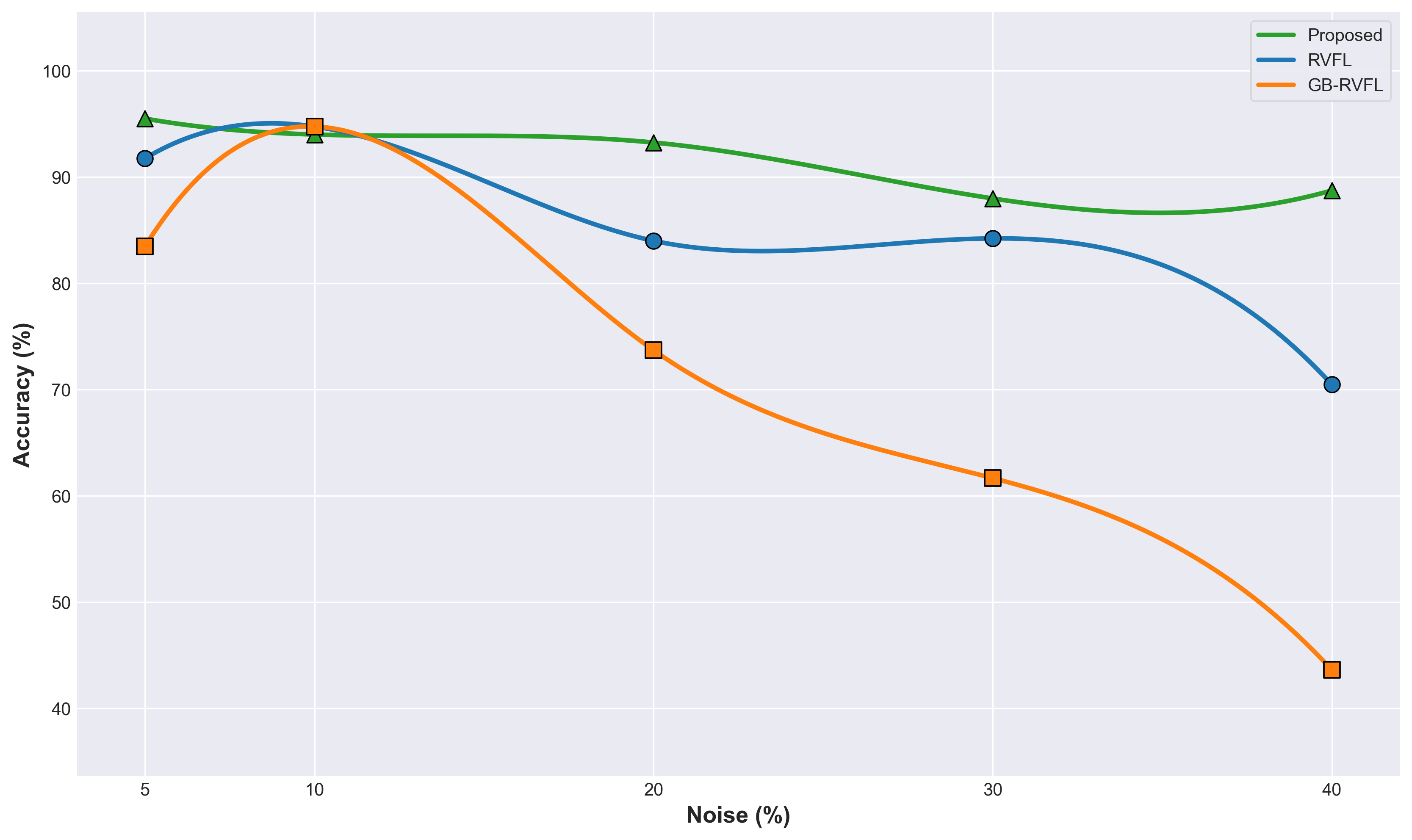}}
\end{minipage}
\par\medskip
\caption{Classification performance plot of models on different levels of label noise.}
\label{noise_plot}
\end{figure*}

\renewcommand{\thefigure}{S.3}
\begin{figure}[htp]
\begin{minipage}{.450\linewidth}
\centering
\subfloat[\label{xi_sensi}]
{\includegraphics[scale=0.25]{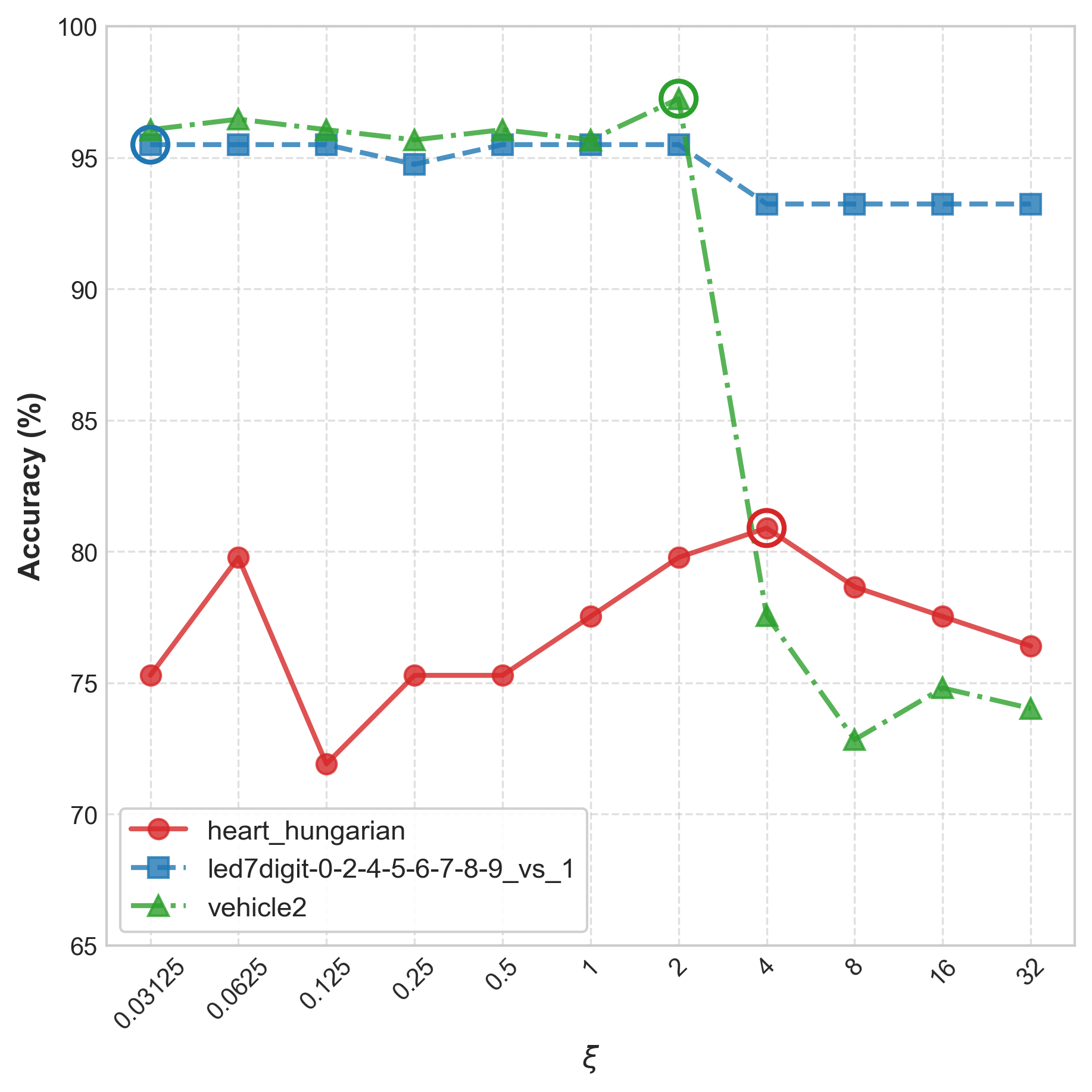}}
\end{minipage}
\begin{minipage}{.450\linewidth}
\centering
\subfloat[\label{rho_sensi}]
{\includegraphics[scale=0.25]{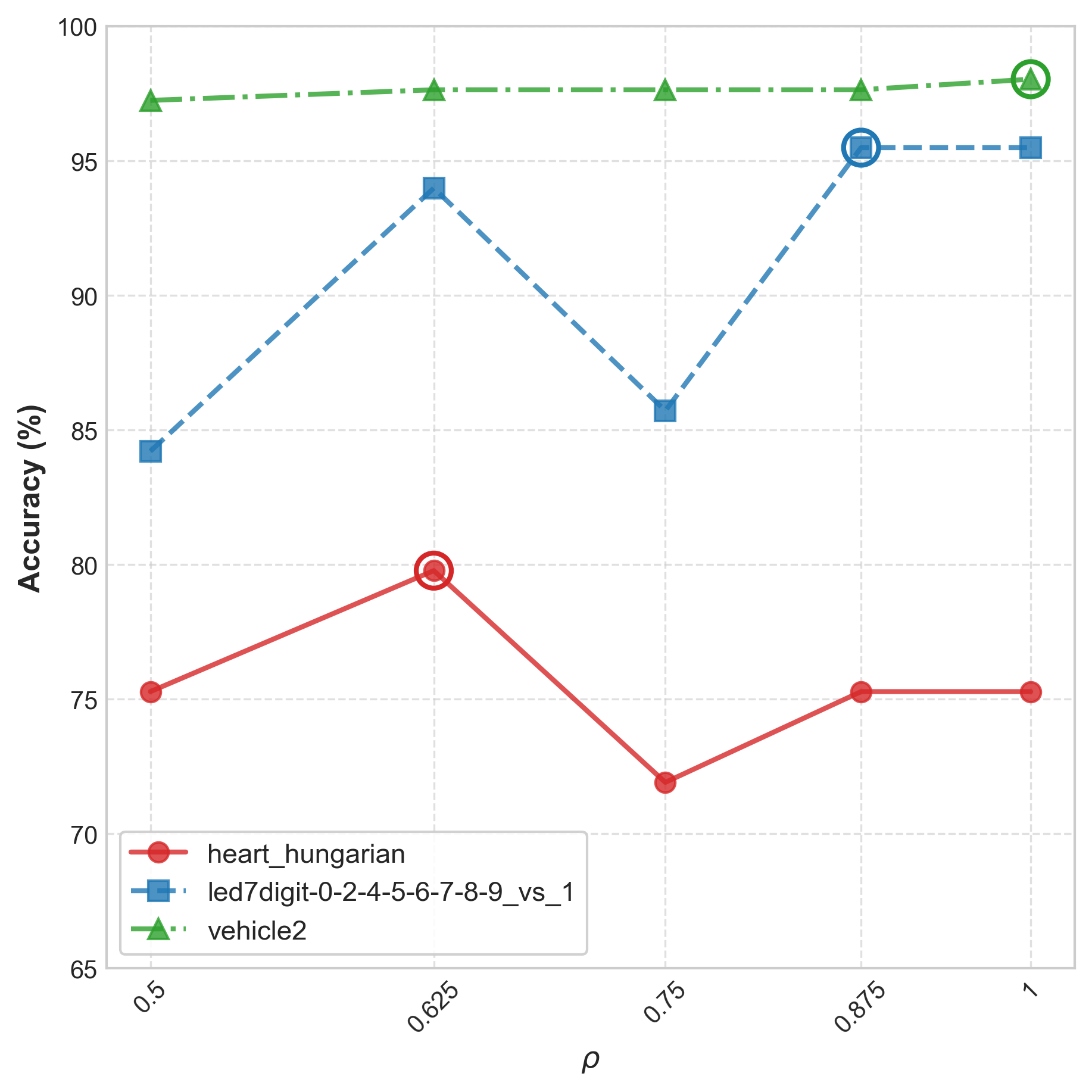}}
\end{minipage}
\par\medskip
\caption{Performance variation of the RoBell-RVFL model with respect to parameters $\xi$ and $\rho$, respectively.}
\label{2d_sensi}
\end{figure}

\renewcommand{\thefigure}{S.4}
\begin{figure*}[!htp]
\centering
\begin{minipage}{0.32\linewidth}
\centering
\subfloat[heart\_hungarian\label{fig:heart_hungarian_3d}]{
\includegraphics[width=\linewidth]{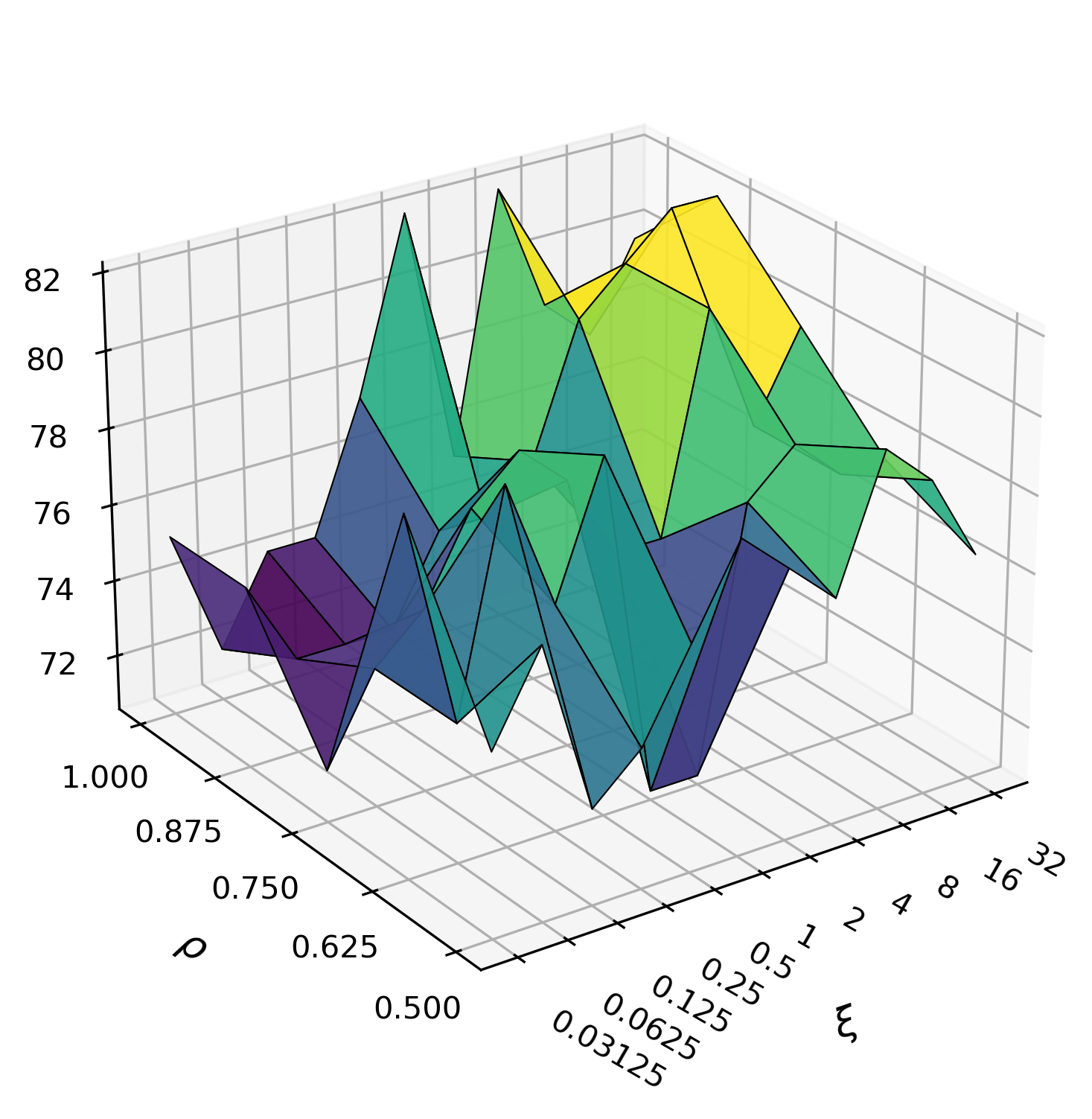}}
\end{minipage}
\hfill
\begin{minipage}{0.32\linewidth}
\centering
\subfloat[vehicle2\label{fig:vehicle2_3d}]{
\includegraphics[width=\linewidth]{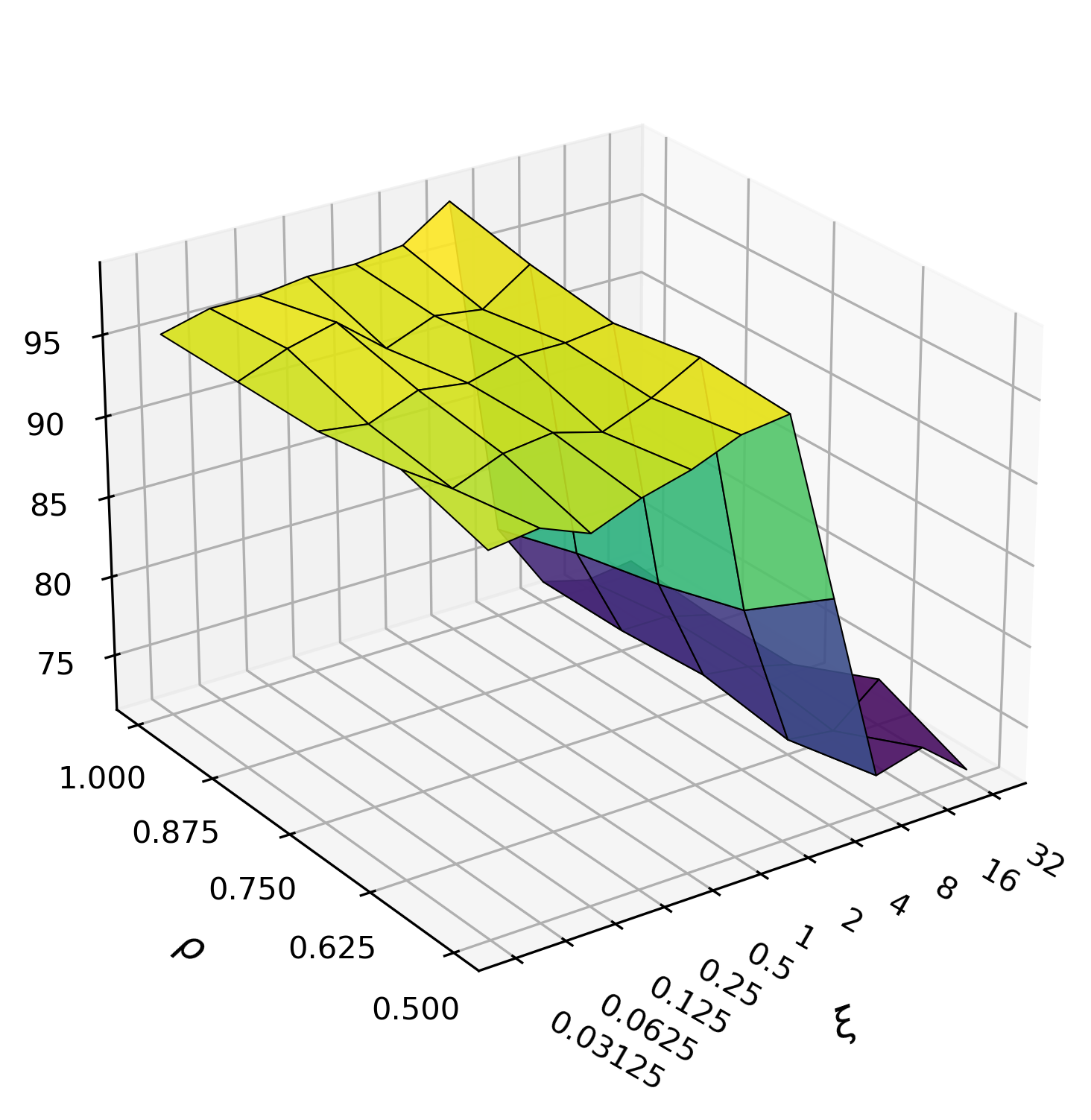}}
\end{minipage}
\hfill
\begin{minipage}{0.32\linewidth}
\centering
\subfloat[led7digit-0-2-4-5-6-7-8-9\_vs\_1\label{fig:led7digit_3d}]{
\includegraphics[width=\linewidth]{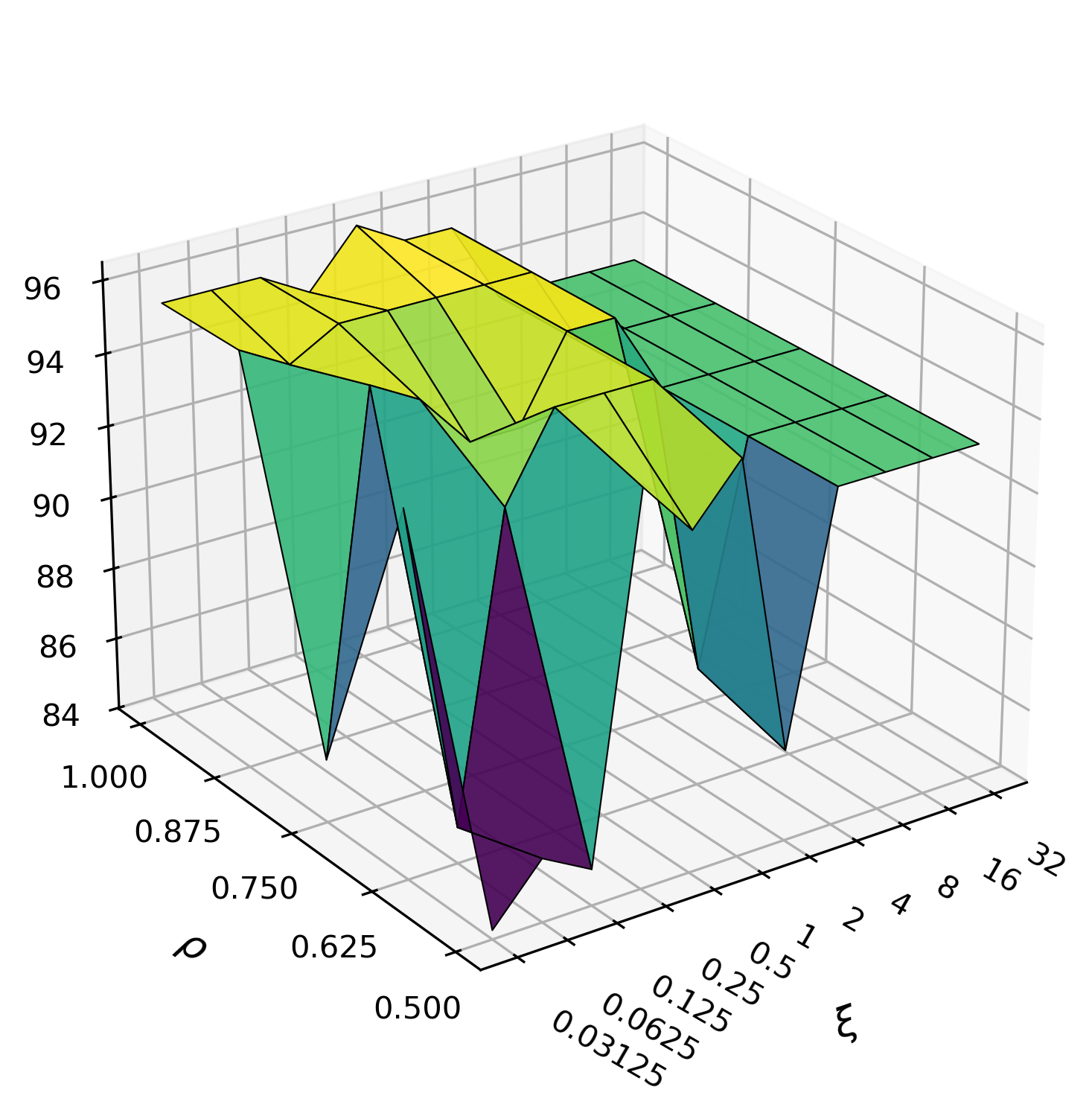}}
\end{minipage}

\caption{Performance variation of the proposed RoBell-RVFL model with respect to parameters $\rho$ and $\xi$ simultaneously.}
\label{fig:3d_sensi}
\end{figure*}

\section*{S.VIII. Sensitivity Analysis}
\label{sensi}
This section investigates the sensitivity of the proposed RoBell-RVFL model to its key hyperparameters, namely the regularization parameter $\xi$ and the parameter $\rho$. To assess the robustness and stability of the proposed framework, sensitivity analyses are conducted on three representative datasets by examining: (i) the effect of $\xi$ on classification accuracy, (ii) the effect of $\rho$ on classification accuracy, and (iii) the joint influence of $\xi$ and $\rho$ on model performance. Each hyperparameter is varied within a predefined range, and its impact on predictive accuracy is systematically analyzed.

\begin{enumerate}[label=(\roman*)]

\item \textbf{Effect of the regularization hyperparameter $\xi$:}

Figure~\ref{xi_sensi} illustrates the influence of the regularization parameter $\xi$ on classification accuracy across different datasets. For the \textit{heart\_hungarian} dataset, accuracy improves as $\xi$ increases up to $4$, beyond which performance begins to degrade. In contrast, the \textit{vehicle2} dataset exhibits a noticeable decline in performance when $\xi > 2$. Meanwhile, the \textit{led7digit-0-2-4-5-6-7-8-9\_vs\_1\_3} dataset remains relatively stable, achieving its peak accuracy at $\xi = 0.03125$. Despite dataset-specific variations, optimal performance across all datasets is consistently observed within the low-to-mid range of $\xi$ values ($\xi \leq 4$). Accordingly, restricting the search space to $\xi \in [0.03125,\,4]$ is sufficient to capture the optimal performance region across diverse datasets.

\item \textbf{Effect of the parameter $\rho$:}

Figure~\ref{rho_sensi} presents the sensitivity of classification accuracy with respect to $\rho$. For most datasets, including \textit{vehicle2}, increasing $\rho$ leads to a gradual improvement in performance. Notably, both the \textit{heart\_hungarian} and \textit{led7digit-0-2-4-5-6-7-8-9\_vs\_1\_3} datasets achieve their maximum accuracy when $\rho \geq 0.625$. These observations indicate that higher values of $\rho$ generally promote stable and improved learning. Consequently, confining the search interval to $\rho \in (0.625,\,1]$ is recommended during hyperparameter tuning.

\item \textbf{Joint effect of $\xi$ and $\rho$:}

Figure~\ref{fig:3d_sensi} depicts the joint sensitivity of testing accuracy with respect to $\xi$ and $\rho$. For the \textit{heart\_hungarian} dataset, the performance surface is irregular, indicating sensitivity to both parameters. High accuracy is achieved when $\rho$ is close to $1$ and $\xi$ assumes moderate values ($\xi \geq 1$), whereas lower $\rho$ values or very small $\xi$ lead to noticeable fluctuations.

In the case of the \textit{vehicle2} dataset, performance is strongly influenced by $\xi$. Optimal accuracy is observed for low-to-moderate $\xi$ values ($\xi \leq 4$) combined with high $\rho$ values. However, excessive $\xi$ results in sharp performance degradation regardless of $\rho$, highlighting sensitivity to over-regularization. In contrast, the \textit{led7digit-0-2-4-5-6-7-8-9\_vs\_1} dataset exhibits a broad and nearly flat performance plateau. Once $\xi \geq 2$, accuracy remains stable across most values of $\rho$, with peak performance occurring around $\rho = 1$ and $\xi \approx 0.5$, indicating strong robustness to joint parameter variations.

Based on the joint sensitivity analysis, dataset-specific tuning ranges can be identified. For \textit{heart\_hungarian}, stable and high accuracy is achieved when $\rho$ is close to $1$ and $\xi$ lies within $[1,\,32]$. For \textit{vehicle2}, optimal performance is obtained with $\xi \in [0.125,\,2]$ and $\rho \approx 1$, while larger $\xi$ values lead to substantial degradation. The \textit{led7digit-0-2-4-5-6-7-8-9\_vs\_1} dataset demonstrates strong robustness across a wide parameter range, particularly when $\rho \in [0.5,\,1]$ and $\xi \geq 0.25$.

Considering all datasets collectively, a practical and robust tuning strategy emerges. Maintaining $\rho$ within the higher range ($\rho \in [0.75,\,1]$) consistently yields stable and superior performance, while low-to-moderate $\xi$ values provide the best balance between accuracy and robustness. Therefore, the combination $\rho \approx 1$ and $\xi \in [0.25,\,2]$ can be regarded as a reliable default configuration for RoBell-RVFL across different datasets.
\end{enumerate}

It is worth noting that model performance may vary depending on the dataset characteristics and application domain. Consequently, careful hyperparameter tuning remains essential for achieving optimal generalization performance.

\renewcommand{\thetable}{S.III}
\begin{table}[htbp]
\centering
\caption{Index of activation functions.}
\label{tab:activation_index}
\renewcommand{\arraystretch}{1.2}
\begin{tabular}{cl}
\hline
\textbf{Index} & \textbf{Activation Functions} \\
\hline
1  & Scaled Exponential Linear Unit (SELU) \\
2  & Rectified Linear Unit (ReLU) \\
3  & Sigmoid \\
4  & Sine (Sin) \\
5  & Hard Limit Transfer Function (Hardlim) \\
6  & Triangular Basis Transfer Function (Tribas) \\
7  & Radial Basis Transfer Function (Radbas) \\
8  & Signum (Sgn) \\
9  & Leaky Rectified Linear Unit (Leaky ReLU) \\
10 & Hyperbolic Tangent Sigmoid Transfer Function (Tansig) \\
\hline
\end{tabular}
\end{table}


\renewcommand{\thetable}{S.IV}
\begin{table*}[!htbp]
\centering
\caption{Best hyperparameters and performance of the proposed model for the experiments on 26 UCI \cite{dua2017uci} and KEEL \cite{derrac2015keel} datasets.}
\label{tab:best_hyperparams}
\renewcommand{\arraystretch}{1}
\resizebox{\textwidth}{!}{%
\begin{tabular}{|l|c|}
\hline
\textbf{Dataset} &
\textbf{(Sensitivity, Specificity, Precision, Recall, F-measure, G-mean, $\gamma$, D, Activation, $\xi$, $\rho$)} \\
\hline
checkerboard\_Data                & (91.2088, 84.6154, 82.1782, 91.2088, 86.4583, 87.8503, 0.01, 163, 6, 2, 1)      \\ \hline
cleve                             & (85.3659, 81.6327, 79.5455, 85.3659, 82.3529, 83.4784, 0, 143, 8, 16, 1)        \\ \hline
cmc                               & (94.332, 32.3077, 63.8356, 94.332, 76.1438, 55.2055, 100000, 183, 9, 2, 0.625)  \\ \hline
congressional\_voting             & (2, 96.2963, 25, 2, 3.7037, 13.8778, 0, 63, 1, 16, 0.625)                       \\ \hline
conn\_bench\_sonar\_mines\_rocks &
  (79.3103, 85.2941, 82.1429, 79.3103, 80.7018, 82.2478, 0.1, 203, 9, 0.125, 0.75) 
   \\
crossplane130                     & (100, 94.7368, 95.2381, 100, 97.561, 97.3329, 0, 63, 4, 0.0625, 0.5)            \\ \hline
crossplane150                     & (88.2353, 96.4286, 93.75, 88.2353, 90.9091, 92.241, 10000, 103, 9, 8, 0.5)      \\ \hline
echocardiogram                    & (53.3333, 100, 100, 53.3333, 69.5652, 73.0297, 0.001, 43, 4, 8, 1)              \\ \hline
ecoli0137vs26                     & (86.6667, 96.2025, 81.25, 86.6667, 83.871, 91.3102, 10, 163, 2, 16, 0.75)       \\ \hline
ecoli2                            & (54.5455, 92.2222, 46.1538, 54.5455, 50, 70.9246, 1, 43, 7, 2, 0.75)            \\ \hline
fertility                         & (33.3333, 92.5926, 33.3333, 33.3333, 33.3333, 55.5556, 1, 143, 5, 8, 0.75)      \\ \hline
haberman                          & (86.8421, 25, 84.6154, 86.8421, 85.7143, 46.5946, 1000, 163, 7, 32, 0.875)      \\ \hline
haberman\_survival                & (25, 86.8421, 28.5714, 25, 26.6667, 46.5946, 1000, 163, 7, 32, 0.875)           \\ \hline
heart\_hungarian                  & (83.871, 70.6897, 60.4651, 83.871, 70.2703, 76.9988, 0, 43, 8, 0.0313, 0.625)   \\ \hline
heart-stat                        & (89.1304, 85.7143, 89.1304, 89.1304, 89.1304, 87.4057, 0, 83, 8, 32, 1)         \\ \hline
iono                              & (95.6522, 81.0811, 90.411, 95.6522, 92.9577, 88.0658, 0.01, 123, 1, 2, 0.625)   \\ \hline
ionosphere                        & (91.25, 69.2308, 90.1235, 91.25, 90.6832, 79.4815, 0.1, 43, 7, 2, 0.5)          \\ \hline
led7digit-0-2-4-5-6-7-8-9\_vs\_1  & (77.7778, 97.5806, 70, 77.7778, 73.6842, 87.1183, 0.0001, 163, 2, 0.5, 1)       \\ \hline
molec\_biol\_promoter             & (55.5556, 78.5714, 76.9231, 55.5556, 64.5161, 66.0687, 1, 143, 2, 1, 1)         \\ \hline
monk1                             & (0, 100, 0, 0, 0, 0, 10, 3, 6, 0.125, 0.875)                                    \\ \hline
monk3                             & (82.0896, 85.0877, 76.3889, 82.0896, 79.1367, 83.5752, 10000, 203, 3, 2, 0.875) \\ \hline
monks\_3                          & (93.5065, 100, 100, 93.5065, 96.6443, 96.6988, 0.1, 163, 2, 1, 1)               \\ \hline
oocytes\_merluccius\_nucleus\_4d  & (84.9246, 65.7407, 82.0388, 84.9246, 83.4568, 74.7195, 1, 183, 9, 2, 0.75)      \\ \hline
vehicle1                          & (81.6667, 84.5361, 62.0253, 81.6667, 70.5036, 83.089, 10000, 123, 9, 2, 0.875)  \\ \hline
vehicle2                          & (94.3662, 98.9071, 97.1014, 94.3662, 95.7143, 96.61, 1, 123, 3, 2, 1)           \\ \hline
yeast-0-2-5-7-9\_vs\_3-6-8        & (71.4286, 98.9051, 86.9565, 71.4286, 78.4314, 84.0515, 100000, 63, 7, 2, 0.5)\\  \hline
\end{tabular}
}
\end{table*}

\end{document}